\documentclass[10pt]{article} 
\usepackage[preprint]{tmlr}

\usepackage{amsmath,amsfonts,bm}

\def\eqref#1{equation~\ref{#1}}

\def\1{\bm{1}}

\DeclareMathAlphabet{\mathsfit}{\encodingdefault}{\sfdefault}{m}{sl}
\SetMathAlphabet{\mathsfit}{bold}{\encodingdefault}{\sfdefault}{bx}{n}

\usepackage{hyperref}
\usepackage{url}

\usepackage{etoc}

\usepackage{booktabs}       
\usepackage{amsfonts}       
\usepackage{nicefrac}       
\usepackage{microtype}      
\usepackage{xcolor}         
\usepackage{float}         
\usepackage{subcaption}
\usepackage{graphicx}

\usepackage{amsmath}
\usepackage{amssymb}
\usepackage{mathtools}
\usepackage{amsthm}
\usepackage{makecell}
\usepackage{bbm}
\usepackage{pifont}
\newcommand{\cmark}{\ding{51}} 
\newcommand{\xmark}{\ding{55}} 

\title{Survival In-Context: Amortized Bayesian Survival Analysis via Prior-Fitted Networks}

\author{%
  \name Dmitrii Seletkov$^{1,2}$ \email dmitrii.seletkov@tum.de \\ \\
  \name Paul Hager$^{1}$ \\ \\
  \name Georgios Kaissis$^{3}$ \\ \\
  \name Rickmer Braren$^{4}$ \\ \\
  \name Daniel Rueckert$^{1,5,6}$ \\ \\
  \name Raphael Rehms$^{1,5}$ \\ \\
  \addr $^{1}$ Chair for AI in Healthcare and Medicine, Technical University of Munich (TUM) and TUM University Hospital, Germany \\
  \addr ${2}$ Institute of Diagnostic and Interventional Radiology, TUM University Hospital, Germany \\
  \addr $^{3}$ Hasso Plattner Institute for Digital Engineering, University of Potsdam, Germany \\
  \addr $^{4}$ Department of Diagnostic and Interventional Radiology and Nuclear Medicine, University Medical Center Hamburg-Eppendorf, Germany \\
  \addr $^{5}$ Munich Center for Machine Learning (MCML), Germany \\
  \addr $^{6}$ Department of Computing, Imperial College London, UK
}

\def\month{MM}  
\def\year{YYYY} 
\def\openreview{\url{https://openreview.net/forum?id=XXXX}} 

\begin{document}
\etocdepthtag.toc{main}

\maketitle

\begin{abstract}
  Survival analysis is crucial for many medical applications, but remains challenging for modern machine learning due to limited data, censoring, and the heterogeneity of tabular covariates. While the prior-fitted paradigm, which relies on pretraining models on large collections of synthetic datasets, has recently facilitated tabular foundation models for classification and regression, its suitability for time-to-event modeling remains unclear. We propose a flexible survival data generation framework that defines a rich survival prior with explicit control over covariates and time-event distributions. Building on this prior, we introduce Survival In-Context (SIC), a prior-fitted in-context learning model for survival analysis that is pretrained exclusively on synthetic data. SIC is trained to approximate Bayesian posterior predictive inference under the synthetic survival prior, enabling individualized survival prediction in a single forward pass, requiring no task-specific training or hyperparameter tuning. Across a broad evaluation on real-world survival datasets, SIC achieves competitive or superior performance compared to classical and deep survival models, particularly in small and medium-sized data regimes, highlighting the promise of a prior-fitted paradigm for survival analysis. The code and pretrained models are available at \url{https://github.com/yayapa/sic}.
\end{abstract}

\section{Introduction}
\label{sec:intro}

In many predictive modeling scenarios, it is not enough to know~\textit{if} something happens, but also~\textit{when}. Survival analysis, also known as time-to-event analysis, models the time until an event of interest, spanning diverse problems such as machine failure, customer churn, and disease diagnosis. However, the target is often only partially observed: devices do not always fail, customers do not always leave, and potential patients may drop out of follow-up in the observed time window. Such cases are censored: the event has not occurred by the end of recording, and it is unknown what happens afterward. Treating time-to-event prediction naively as regression (e.g., discarding censored samples) wastes information and introduces bias towards earlier events. Survival analysis uses both complete and censored outcomes, yielding risk estimates that evolve over time. This challenging problem is especially important for clinical decision-making, where models for disease manifestation~\citep{seletkov26}, relapse~\citep{zupan00}, and survival outcomes~\citep{tesfay21} must be able to model how a patient's risk changes.

In practice, survival analysis is most commonly performed on tabular data, particularly in healthcare, where each patient is often represented by heterogeneous covariates spanning demographics, measurements, and interventions. Despite its prevalence, tabular data poses challenges that fundamentally distinguish it from unstructured modalities such as text or images~\citep{bommasani21}. 
In contrast to language modeling, where tokens share a consistent semantic meaning across datasets, a tabular value is a number whose meaning comes from its column name (and units), e.g., 120 for systolic blood pressure (mmHg) and temperature (${}^{\circ}\mathrm{C}$).
This heterogeneity has historically favored classical task-specific survival approaches such as Cox Proportional Hazards~\citep{cox72} or Accelerated Failure Time models~\citep{prentice79}.

These problems have also long held back tabular foundation models in classification and regression~\citep{hollmann25}, in contrast to the success of foundation models in language~\citep{brown20} and vision~\citep{radford21}. Recent advances attempt to address this gap by introducing the prior-fitted paradigm of TabPFN models~\citep{hollmann23, hollmann25}. TabPFN is trained across millions of synthetic tabular datasets to act as a learning algorithm rather than a single-task predictor: given a new dataset, TabPFN uses in-context learning to infer dataset-specific structure and produce predictions in a single forward pass. 
From a Bayesian perspective, this can be viewed as approximating a posterior predictive distribution over the synthetic data prior used during pretraining: the observed context dataset conditions the prediction, while the pretrained network amortizes the inference procedure.
This helps it outperform strongly tuned baselines on tabular datasets while being orders of magnitude faster at deployment, effectively folding model selection and hyperparameter search into pretraining.
This paradigm is particularly interesting for survival analysis, where labeled data are often scarce, and privacy constraints often prevent assembling large cohorts, since no training needs to be done on the actual data.

Recent work shows further successes~\citep{mueller25} of the prior-fitted paradigm in scalability and efficiency~\citep{qu25, liu25, ma25} and across application tasks, including causal effect estimation~\citep{robertson25}, outlier detection~\citep{shen25}, and time-series forecasting~\citep{hoo24}. However, extending this paradigm to survival analysis is challenging, since censoring and time-dependent risk require modeling distributions over event times rather than point labels. In this work, we bring prior-fitted in-context learning to survival analysis and evaluate it on heterogeneous public cohorts, reducing reliance on dataset-specific evidence. 
To summarize our contributions:

\begin{itemize}
    \item We introduce a novel framework for generating synthetic survival data, with explicit control over covariate and survival time distributions.
    \item We propose a prior-fitted in-context learning model for survival analysis, called Survival In-Context (SIC), which is pretrained solely on synthetic data. The model produces predictions in a single forward pass and requires no hyperparameter tuning, unlike classical survival models. This makes it particularly attractive for small- to medium-sized datasets.
    \item We compile a diverse set of 58 publicly available real-world survival datasets from previous work and major libraries to enable faithful comparisons across heterogeneous settings, evaluating against 10 established baselines with extensive hyperparameter tuning.
  
\end{itemize}

\section{Theoretical Background and Related Work}
\label{sec:background}
 \subsection{Tabular Foundation Models and Prior Fitted Networks} 
 \label{sec:background:tfm}
 By leveraging multiple datasets and learning common representations, tabular foundation models promise to both improve performance and reduce implementation time through model reuse. Recent transformer-based models tackle learning across tables by constructing joint representations from a varying number of inputs~\citep{zhu23}, utilizing LLMs as tabular learners~\citep{gardner24}, or adding column semantics to the input data~\citep{wang22, kim25}.

The prior-fitted paradigm in TabPFN~\citep{mueller22, hollmann23} advances tabular foundation models through two key ingredients: pretraining on a large and expressive prior, i.e., synthetic datasets designed to match the target domain, and in-context learning to approximate the Bayesian posterior predictive distribution (PPD) induced by this prior. Subsequent work has improved PFNs through richer prior generators and architecture of the model~\citep{hollmann25, grinsztajn25}, refined the context selection strategy~\citep{thomas24, koshil24}, and extended context size, allowing to utilize a larger amount of training data ~\citep{qu25, liu25}, showing effectiveness in classification and regression problems~\citep{erickson25}.

\citet{nagler23} backs the PFN methodology with a rigorous statistical analysis. If we take a Bayesian generative perspective, the context (or training) data $\mathcal D_n^\text{train}=\{(x_i,y_i)\}_{i=1}^{n}$ are sampled from a realization of an infinite-dimensional random variable $p \in \mathcal P$ with a prior $\Pi$ over possible distributions. Thus, $\Pi$ encodes assumptions of our belief about the possible distributions and input-output relationships the model is expected to encounter. A PFN $q_\theta$ is trained to approximate the PPD for a new query point $x^*$, given the context data $\mathcal D_n^\text{train}$: 
 
\begin{equation}
\label{eq:ppd}
 \pi(y \mid x^\ast,\mathcal D_n^\text{train})
    =
    \int p(y \mid x^\ast)\, d\Pi(p \mid \mathcal D_n^\text{train})
\end{equation}

This is the expectation over the posterior distribution of models given the training data. ~\citet{mueller22} propose to learn a PFN model $q_{\theta^*}$ that yields an approximation for this PPD through optimization:

\begin{equation}\label{eq.objectivePFNs}
    \hat \theta
    =
    \arg\max_{\theta} \mathbb{E}_{\Pi_N}
    \mathbb{E}_{\Pi}
    \left[
        \log q_{\theta}(Y \mid X,\mathcal D)
    \right]
\end{equation}
where $\Pi_N$ is a probability distribution over the sample size $N$. The expectation over $N$ and dataset types is approximated by Monte Carlo sampling using synthetic datasets drawn from the prior $\Pi$ with sample size $N$. Once trained, $q_\theta$ amortizes approximate Bayesian posterior predictive inference, enabling prediction on a new dataset in a single forward pass without task-specific parameter updates or hyperparameter tuning. Consequently, the prior $\Pi$ is the central modeling assumption, as it defines the space of data-generating mechanisms over which the PFN learns to infer. Extending PFNs to survival analysis, therefore, requires designing a survival-specific prior over censored time-to-event datasets. Our work builds on previous work in the prior-fitted paradigm, focusing on a new prior tailored to survival analysis.

\begin{figure}[t]
  \begin{center}
    \centerline{\includegraphics[width=\textwidth]{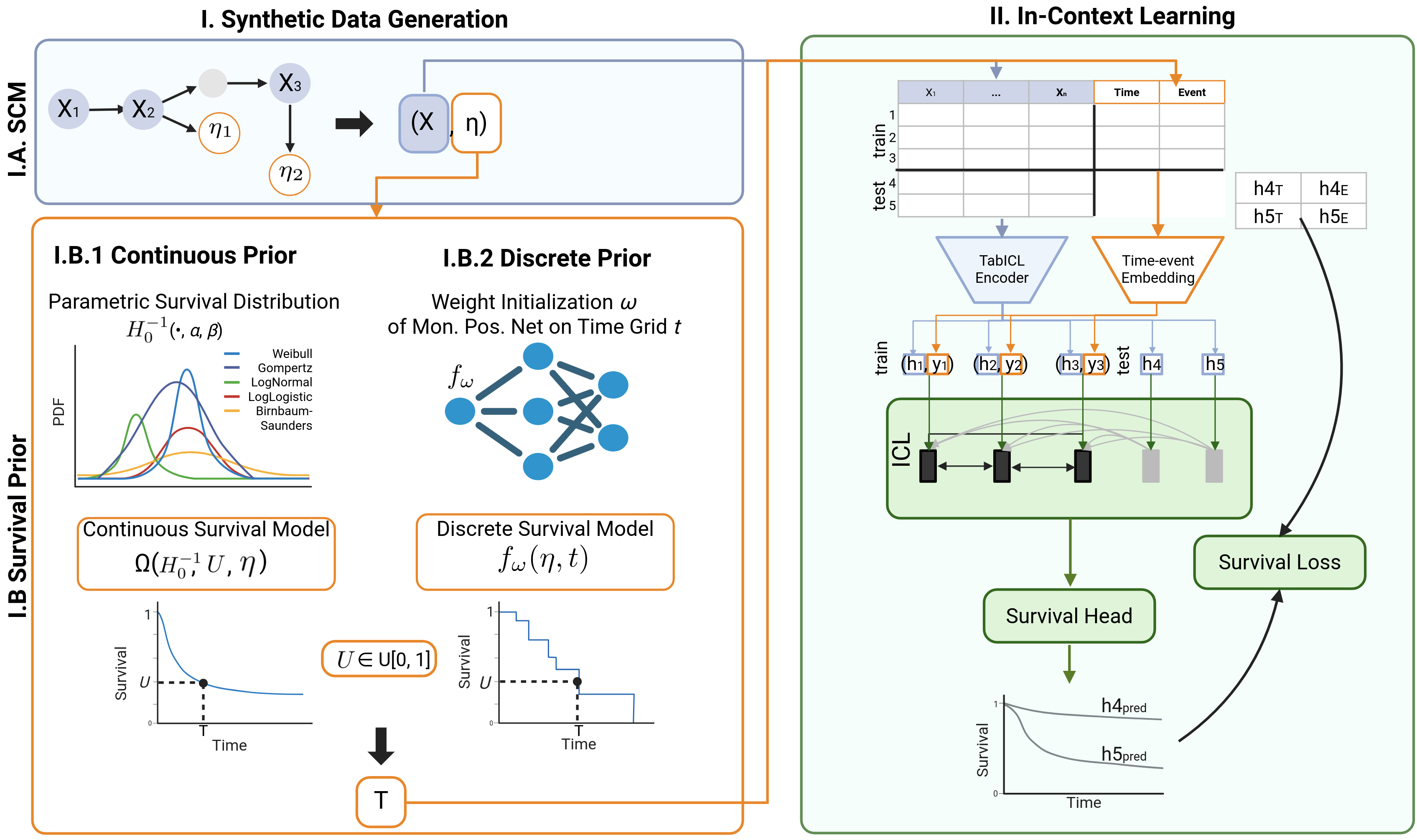}}
    \caption{Survival In-context (SIC) method contains two stages. (I) Synthetic Data Generation using Structural Causal Models (SCMs) for the generation of (I.A) covariates X and the associated label nodes $\eta$, followed by (I.B) two proposed survival priors to generate time-event labels: continuous (I.B.1) with the survival function parametrization $H_0^{-1}(*, \alpha, \beta)$ and discrete (I.B.2) with Monotonic Positive Network~\citep{Rindt21} $f_\omega$. (II) In-context learning on synthetic data with a specialized time-event embedding and survival head and loss. Created in \url{https://BioRender.com}.}
    \label{fig:sic}
  \end{center}
\end{figure}

\subsection{Survival Analyis}
Survival analysis~\citep{kleinbaum12} models the data as a set of $\{x_i, t_i, e_i\}$, with $x_i\in\mathbb{R}^d$ the $d$-dimensional input, $e_i \in \{0, 1\}$ indicates the occurrence of event ($e_i=1$, e.g., death) or right-censoring ($e_i=0$, e.g., observation time end), $t_i \in \mathbb{R^+}$ the time until event or censoring for the subject $i$. The prediction target is a monotonically decreasing survival function $S(t|x)\in [0, 1]$ that indicates the probability that an individual survives beyond time $t$. The other estimated targets are the hazard function $h(t|X)$ describing the instantaneous rate of event conditioned on surviving up to time $t$, and the cumulative hazard function $H(t|x)$ with the following relations to the survival function:

\begin{equation}
    S(t|x) = \mathbb{P}(T>t|X=x) = 1 - \int_{0}^t f(s)ds = exp(-H(t|x)) = exp(-\int_{0}^{t}h(u|x)du)
\end{equation}

Survival analysis methods can be categorized by whether they model time as continuous or discrete, which in turn determines the assumptions they impose on the hazard or survival function. 

To model time continuously, models impose structural assumptions between covariates and the event risk. Cox Proportional Hazards (CoxPH) model~\citep{cox72} is semi-parametric and assumes the hazard of one individual is proportional to the hazard of any other individual. When this assumption fails, the accelerated failure time (AFT) model~\citep{prentice79} offers an alternative. This assumes the basic shape of the survival function could be stretched along the time axis to accelerate or decelerate the event occurrence. 

However, when domain knowledge suggests a particular prior, assuming the distribution of events can be advantageous, especially in limited data scenarios. Yet, this is strongly application-dependent, e.g., Weibull-CoxPH for gastric cancer~\citep{zhu11},  Gompertz-CoxPH for aging mortality~\citep{lenart10}, Log-normal-CoxPH for gallbladder cancer~\citep{wang10}, Log-logistic-CoxPH~\citep{kumar20} for lung cancer, Birnbaum-Saunders-CoxPH~\citep{leao17} for leukemia. As an extension to a wide array of statistical methodologies, the deep learning variants DeepSurv~\citep{katzman18} and DeepAFT~\citep{norman24} generalize CoxPH and AFT by replacing linear predictors with flexible neural network-based function approximators. Deep Extended Hazard (DeepEH)~\citep{zhong21} further unifies PH and AFT formulations under the extended hazard assumption~\citep{amoli87} as special cases. 

An alternative line of work models time discretely. Kaplan-Meier (KM) estimator~\citep{km58} is non-parametric and makes no assumptions about the underlying event-time distribution or covariate effects, but only describes a population-level survival. This has been further extended to incorporate covariates in survival function estimation~\citep{beran81, chen19, chen20, chen23, chen24}. Another discrete-time method, DeepHit~\citep{lee18}, directly estimates the distribution of events in discrete time, using a neural network trained with a likelihood loss and a ranking loss. While discrete-time methods are more flexible, they are known to be sensitive to the discretization grid choice. 

To internalize time discretization and interpolation steps, the cumulative hazard can be obtained by solving a neural ODE~\citep{danks22, tang22, moon22} or directly through monotone positive neural networks~\citep{Rindt21, jeanselme23}. 

\paragraph{Concurrent Work on Prior-fitted Networks for Survival Analysis.} Earlier ~\citet{craig21} proposed Survival Stacking, formulating survival analysis as a sequence of binary classification problems, modeling discrete-time hazards at unique event times, and enabling the use of an arbitrary binary classifier. Building on this idea, \citet{kim26r} instead models cumulative failure probabilities at a fixed discrete-time grid and applies off-the-shelf classification PFNs. More closely related to our work, both SurvPFN~\citep{boehm26} and SurvivalPFN~\citep{qi26} introduce a survival prior. While SurvPFN relies on the Weibull distribution, SurvivalPFN uses two mechanisms: random monotone Bernstein maps and mixtures of Weibull and Lognormal distributions. We include SurvivalPFN, SurvPFN, and ~\citet{kim26r} in our analysis.

Additional related work in synthetic data generation for Survival Analysis is provided in Appendix~\ref{app:relatedwork}.

\section{Survival In-context}
\label{sec:method}

Figure~\ref{fig:sic} overviews the proposed Survival In-Context (SIC) framework that consists of two main components: synthesizing data using the survival prior in~\ref{sic:prior} and the in-context learning (ICL) architecture of our survival analysis model in~\ref{sic:icl}.

\subsection{Synthetic Data Generation}
\label{sic:prior}
\subsubsection{Covariate Generation.} Survival in-context (SIC) model is trained solely on synthetic data. For the data synthesis, we employ the Structural Causal Models (SCMs)~\citep{pearl09} proposed for data generation in TabPFN~\citep{hollmann22} and later enhanced in TabICL~\citep{qu25}. We first sample a directed acyclic graph (DAG), where nodes correspond to the variables and edges to MLP or tree-based relations. This allows a wide range of potentially complex and non-linear dependencies. Each feature $c$ is then modeled as a function $g$ of its parent variables $Pa(c)$ in the graph, with added independent noise $~\epsilon \sim\mathcal{N}(0, \sigma^2I)$, i.e., $c = g (Pa(c)) + \epsilon$ . The target and train features are selected randomly. In contrast to the classification and regression data synthesis, we generate multiple target variables $\{\eta_i\}_{i=1}^K$ depending on the prior, which results in the synthetic dataset $(X, \eta)$. For the survival prior, we apply a z-score normalization to $\eta$.

\subsubsection{Survival Prior}
To synthesize valid survival analysis datasets, we generate survival curves explicitly conditional on the covariates $X$ via the targets $\eta$, and use an inverse sampling transform to obtain the corresponding time-to-event labels. To capture diverse survival mechanisms, we propose two different survival priors. First, a continuous parametric prior that uses survival functions from known survival models and allows deriving time-to-event labels in closed form. The prior is motivated by classical statistical survival models. Second, we propose a discrete prior that constructs survival curves on a time grid using Monotonic Positive Networks~\citep{Rindt21}. This approach is agnostic about the functional form of survival curves, but it requires time discretization to obtain time-to-event labels. Further, we present both priors and how they complement each other.

\paragraph{Continuous Prior.} To cover a wide variety of survival assumption modeling, such as Proportional Hazard (PH), Accelerated Failure Time (AFT), and Accelerated Hazard (AH), we employ the Extended Hazard (EH)~\citep{amoli87} as a well-behaved, yet very flexible model:

\begin{equation}
    h(t|x)=h_0\left(t e^{\eta_1}\right) e^{\eta_2}
\end{equation}

This allows modeling a broader range of survival curves, e.g., beyond the non-crossing constraints typical of PH. Setting $\eta_1=0$ results in the standard PH, $\eta_1=\eta_2$ in the AFT, and $\eta_2=0$ in the AH assumption. We derive the time-to-event variable under EH assumption (see Appendix~\ref{app:T}):

\begin{equation}
    T=e^{-\eta_1}\, H_0^{-1}\!\left(e^{\eta_1-\eta_2}(-\log U)\right)
\end{equation}
where $(\eta_1, \eta_2)$ are target variables generated by SCMs, $U\in[0, 1]$ is a survival quantile, and $H_0^{-1}$ is an inverse cumulatative hazard function. This may follow the functional form of a Weibull, Lognormal, Loglogistic, Gompertz, or Birnbaum-Saunders distributions, parameterized by scale $\alpha$ and shape $\beta$, as the superset of the established baseline survival distributions in the closed form~\citep{survstan}. Note that, e.g., typical for survival analysis, Exponential and Rayleigh distributions are the special cases of Weibull distributions, and Gamma and Generalized Gamma do not have a closed form and, thus, cannot be directly used in the synthetic data generation. The exact formulae for the inverse cumulative hazard distributions are reported in Appendix~\ref{app:H0-1}.

\textit{Censoring.} To simulate right-censored data (we refer to the terminology in~\citet{booksa}), we draw censoring times $C$ using the same sampling mechanism as for event times with the risk scores $(\eta^{C}_1,\eta^{C}_2)$ that are either set to $0$ (marginally independent censoring, $C\perp (T,X)$) or produced by a separate SCM whose only input is the observed feature matrix $X$ (feature-dependent censoring,
$C\not\perp X$).
To simulate administrative censoring (end of study due to limited follow-up time or a limited number of events), we additionally compute the cutoff time $T_{adminType1}$ at the random quantile of the time or $T_{adminType2}$ at the random quantile of events and truncate observation times at these cutoffs, resulting in $T'_{C} = min(C, T_{adminType1}, T_{adminType2})$. The final observed times are $T_{obs} = min(T,T'_{C})$ with $e=\mathbf{1}\{T<T'_{C}\}$. We enforce in the SCM construction that feature-dependent censoring does not share any unobserved confounders in the SCM, making it non-informative, $C \perp T \mid X$. Therefore, the censoring can be identified from $(X,T_{obs},e)$. We additionally verify these properties in Appendix~\ref{app:posthoccheck} on generated datasets in the ablation study. 

\paragraph{Discrete Prior.}
To complement the closed-form continuous prior with a more flexible family of survival distributions, we define a discrete prior based on randomly initialized positive monotone neural networks $f_{\omega}$ with $W^{(\ell)} > 0$, where:
\begin{equation}
    f_\omega(\eta, t)
    =
    \sigma\!\left(
        W^{(L)} h^{(L-1)} + b^{(L)}
    \right),
    \qquad
    h^{(\ell)}
    =
    \tanh\!\left(
        W^{(\ell)} h^{(\ell-1)} + b^{(\ell)}
    \right),
    \
    h^{(0)} = (\eta, t),
\end{equation}

Because all weights are positive and both $tanh$ and $\sigma$ are monotone increasing functions, $f_\omega(\eta, t)$ is monotone increasing in t. To obtain a valid survival curve on the full normalized time range, we anchor the endpoints. In particular, we normalize the network output such that the survival curve satisfies $S(t=0, \eta) = 1$ and $S(t=1|\eta) = 0$ (see details in Appendix~\ref{app:surv_prior:anchor}). 

We construct a survival curve by evaluating $S(t|\eta) = \{S_k\}_{k=0}^K$ on the normalized increasing time grid $\{t_k\}_{k=0}^K \in [0, 1]$ with $K$ bins. We sample a survival quantile $U \sim \mathcal{U}[0, 1]$ and solve $S(T|\eta) = U$. Since the survival curve is evaluated on the grid, we find the interval $[t_k, t_{k+1}]$ so that $S_k > U > S_{k+1}$. Within this interval, we assume that the instantaneous hazard is constant. This corresponds to exponential interpolation between the two neighboring survival values, resulting in the event time:
\begin{equation}
     T = t_k + (t_{k+1} -t_k) \frac{log(S_k/U)}{log(S_k/S_{k+1})}
\end{equation}

\textit{Weight initialization.} To obtain diverse but numerically stable survival curves, we initialize the
positive weights using a Gamma distribution $\Gamma(k, \theta)$, where $k>0$ is a shape and $\theta>0$ is a scale parameter. To control the pre-activation variance, we parameterize the weight scale in terms of the number of input neurons (fan-in) $n$ and gain $\alpha$, similarly to the Xavier initialization~\citep{glorot10}, which keeps the variance approximately stable across layers. The resulting weight initialization:

 \begin{equation}
     W \sim \Gamma(k, \theta = \frac{\alpha}{\sqrt{n k(k+1)}}) 
 \end{equation}

 This reparameterization allows us to control the Gamma distribution through two interpretable parameters: $\alpha$ sets the overall scale, and $k$ controls heterogeneity among positive weights. Larger $\alpha$ produces sharper survival curves, whereas smaller $\alpha$ produces near-diagonal curves. Smaller $k$ induces sparse, heterogeneous survival families, while larger $k$ produces smoother, more homogeneous ones. Appendix~\ref{app:surv_prior:weightinit} provides the full derivation and visualizations of the proposed parameters.
 
\textit{First layer initialization.}
Positive weights introduce a nonzero mean shift in the first pre-activation $z_j^{(1)}$. For a first-layer unit with input $(\eta,t)$ and zero bias, $\mathbb{E}[\eta]=0$ and $\mathbb{E}[t]=0.5$. Therefore, $\mathbb{E}[z^1] = \mathbb{E}[W_\eta \eta + W_t t] = \frac{1}{2} \mathbb{E}[W_t]$. Thus, without correction, the input to $\tanh$ is shifted away from zero, which
can push the activation into saturation and reduce the diversity of the
generated survival curves. We therefore introduce two first-layer controls: a time-centering bias parameter
$\mu_t$ and a covariate-effect scale $\lambda_\eta$. For a hidden unit $j$, the
first pre-activation with a bias is defined as:

\begin{equation}
    z_j^{(1)}
    =
    \lambda_\eta W_{\eta,j}^{(1)} \eta
    +
    W_{t,j}^{(1)}(t-\mu_t).
\end{equation}
The parameter $\mu_t$ is a reference time location used to shift the time axis; choosing $\mu_t=0.5$ centers the network around the midpoint of the normalized grid, whereas sampling $\mu_t$ produces different temporal regimes. $\lambda$ controls the relative strength of covariate-dependent variation.

\textit{Censoring.} Similarly to continuous prior, we generate right-censored data with the same sampling mechanism as event times, using a separate monotone network and risk scores $\eta^{C}$ to introduce conditionally independent non-informative censoring ($C \perp T \mid X$) or a time-only monotone network that does not depend on covariates to introduce marginally independent censoring ($C \perp (T, X)$). We rescale both event and censoring times by $t_{\max}$ and apply administrative censoring as in the continuous prior (more details are provided in Appendix~\ref{app:surv_prior:censor}).

Together, $(\alpha, k, \mu_t,\lambda)$, define a prior over survival curve shapes. Examples of concrete survival curves and how they depend on the proposed parameters are shown in Appendix~\ref{app:surv_prior:params}.

\begin{figure}
    \centering
    \includegraphics[width=\linewidth]{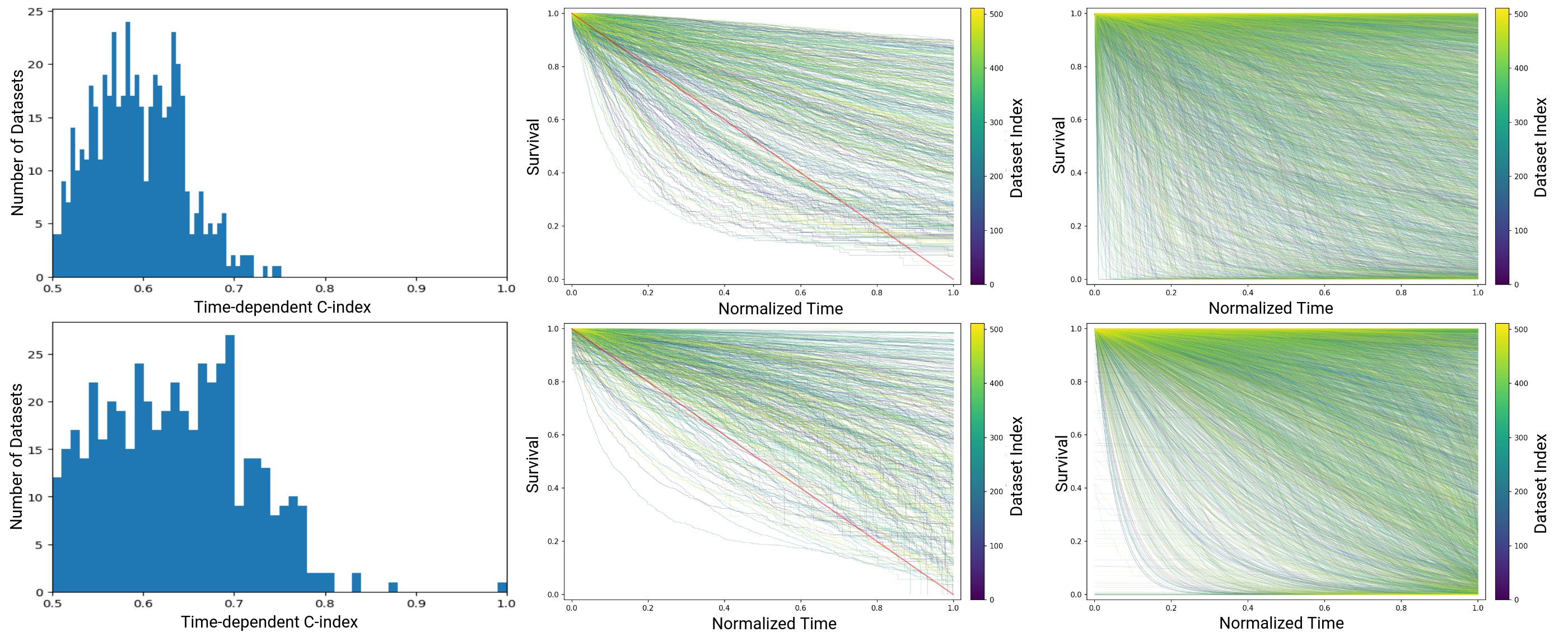}

\caption{
Visualization of a batch of 512 generated
datasets using SCM $\eta_1=X_1 +X_2 \cdot X_3$ and $\eta_2 = X_1 \cdot X_2$ for continuous (upper) and discrete (lower) survival priors: time-dependent C-index estimated from fit CoxPH models (left), Kaplan–Meier curves (middle), and the underlying event survival curves on the normalized time scale. Both proposed synthetic data generation methods demonstrate the ability to generate diverse outcome complexity (left), dataset-level survival patterns (middle), and individualized survival curves (right).}
\label{fig:prior_ablation}
\end{figure}

\paragraph{Mixed Survival Prior.} Finally, to examine the synthesized datasets, we fit the CoxPH models on a random batch of 512 datasets to calculate C-statistics and visualize survival and Kaplan-Meier curves across datasets. The resulting plots are shown in Figure~\ref{fig:prior_ablation}. This analysis demonstrates the ability of our data synthesis process to generate a diverse spectrum of complexity and variety in the survival datasets. We include both types of prior in the final SIC training.

\subsection{In-context Learning}
\label{sic:icl}

\paragraph{Model Architecture.} We employ the TabICL~\citep{qu25} model as the in-context learning architecture. The \textit{TabICL Encoder} consists of the original column-wise inter-sample embedding and row-wise inter-feature interaction, producing $4\times\text{[CLS]}$ tokens $h_i$ per sample. To incorporate survival labels, we use a \textit{Time-event Embedding}: the event $e_i$ is embedded via a linear layer applied to one-hot encoding, and the event time $t_i$ is embedded using a standard multilayer perceptron (MLP). We combine these embeddings multiplicatively and add them to the sample representation: $h_i = h_i + Linear(OneHot(e_i))*MLP(t_i)$. 

We select the TabICL~\citep{qu25} architecture due to its favorable computational scaling and practical usability: for $n$ samples and $m$ features, TabICL has complexity $\mathcal{O}(m^2n + n^2)$ compared to $\mathcal{O}(m^2n + n^2m)$ of the TabPFN. This allows TabICL to perform inference up to 10 times faster and handle up to the recommended 100K samples, compared to TabPFNv2, which supports only 10K samples. In addition, TabICL’s prior generation and training pipeline is publicly available, whereas TabPFNv2's prior generation is not.

\paragraph{Survival Head and Loss.} 
To produce survival outputs, we employ SuMo++~\citep{sumopp}, a modification of SuMo~\citep{Rindt21}, which, in addition to monotonicity in $t$, guarantees $S(0|x) = 1$ by construction. Given an ICL embedding \(q\) of the query subject, SuMo++ models the cumulative hazard $H$ with a monotone positive network \(M_+\) (monotone in \(t\), unconstrained in \(q\)): $S(t\mid q) = \exp\bigl(-H(t\mid q)\bigr) \text{ with } H(t\mid q)=M_+(t,q)-M_+(0,q)
$

This directly parameterizes a continuous-time survival function without imposing a hazard assumption (e.g., CoxPH~\citep{cox72}) or time discretization (e.g., DeepHit~\citep{lee18}). Alternative survival heads are further discussed in Section~\ref{res:ablation}.

Consequently, as the \textit{Survival Loss}, we use the right-censored negative
log-likelihood, which is a proper scoring rule under non-informative censoring~\citep{Rindt21}:
\begin{equation}
\label{eq:nll}
  \mathcal{L}_{\mathrm{NLL}}
  =
  -\frac{1}{N}
  \sum_{i=1}^{N}
  \Bigl[
    e_i\,\log f(T_i\mid q_i)
    +
    (1-e_i)\,\log S(T_i\mid q_i)
  \Bigr].
\end{equation}
where $q_i$ denotes the \textit{ICL} embedding of query subject \(i\), with event $e_i$, time $T_i$, and event-time density $f$.

\paragraph{Large-scale SIC training on Prior.}
Similar to the pretraining strategy of TabICL, inspired by
LLMs that first learn on shorter sentences before moving to longer ones, we gradually increase the number of samples, which is the main complexity factor of the prior~\citep{nagler23}. Therefore, we train the model in a 2-stage procedure. Details are provided in Appendix~\ref{app:training}. 

\paragraph{Bayesian Inference.}
After training with Equation~\ref{eq:nll}, SIC predicts the survival probability for arbitrary $t$ and query point $x^\ast$ with the context $\mathcal D^{\text{train}}=\{(x_i,t_i,e_i)\}_{i=1}^n$,  via the survival PPD:
\begin{equation}\label{eq:survppd}
  S(t\mid x^\ast,\mathcal D^{\text{train}})=\int S_p(t\mid x^\ast)\,d\Pi(p\mid\mathcal D^{\text{train}}).
\end{equation}
where $S_p$ and $f_p$ denote the survival function and density of the event time under the data-generating distribution $p$ drawn from the survival prior $\Pi$, introduced in Section~\ref{sic:prior}). Under non-informative censoring, the posterior splits into the prior and the censored likelihood:

\begin{equation}
    \label{eq:survposterior}
    d\Pi(p\mid\mathcal D^{\text{train}})\propto\prod_{i=1}^n f_p(t_i\mid x_i)^{e_i}\,S_p(t_i\mid x_i)^{1-e_i}\,d\Pi(p),
\end{equation}

where events contribute their density and censored subjects their survival probability, so censored data inform the posterior instead of being discarded. Since $\mathcal L_{\mathrm{NLL}}$ is a proper scoring rule, its expected minimizer under $\Pi$ is this PPD wherever subjects remain at risk. This motivates a necessity of dedicated survival prior (continuous and discrete) to represent realistic prior beliefs.

\begin{figure*}[t]
  \begin{center}
    \centerline{\includegraphics[width=\textwidth]{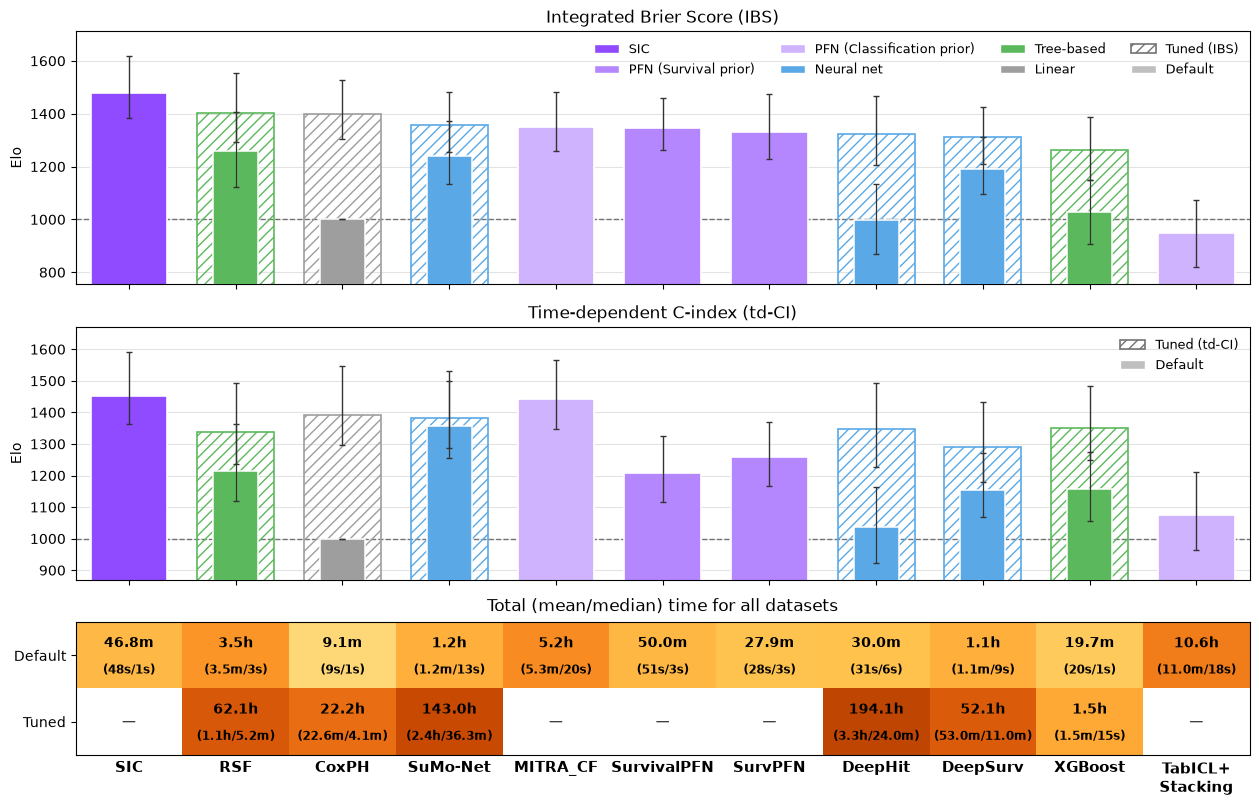}}
    \caption{Performance in Elo rating for Integrated Brier Score (IBS), Time-dependent C-index (td-CI), and total (mean/median) time across 58 datasets spanning 100 to 280,845 samples and 3 to 160 features. The proposed Survival In-context (SIC) method is compared to the classical linear (CoxPH~\citep{cox72}), neural net (SuMo-Net~\citep{Rindt21}, DeepHit~\citep{lee18}, DeepSurv~\citep{katzman18}), tree-based (RSF~\citep{ishawaran08}, XGB~\citep{chen16}) baselines; naive Prior-fitted Network (PFN) TabICL~\citep{qu25} + Stacking~\citep{craig21} baseline and the concurrent PFN baselines MITRA\_CF~\citep{kim26r} with a classification prior, SurvPFN~\citep{boehm26}, and SurvivalPFN~\citep{qi26} with a survival prior. SIC achieves the highest Elo rating requiring no hyperparameter tuning, while maintaining one of the lowest total computation times. Elo ratings are calibrated to CoxPH (default), fixed at 1000. The classical baselines are tuned with 50 trials for IBS in the IBS panel and for td-CI in the td-CI panel. Tuned total time is measured only for IBS-optimized runs.}
    \label{fig:results}
  \end{center}
\end{figure*}

\section{Experiments}
\subsection{Benchmark}
\label{sec:exp}

\paragraph{Datasets.}
Analyzing previous work~\citep{lee18, katzman18, napgal21, zhong21, chen24book, wiegrebe24, knottenbelt25} and based on SurvSet~\citep{survset}, we compiled 58 publicly accessible datasets, ranging from 100 to 280,845 samples, 3 to 160 raw features, and 0.066 to 0.945 censoring rates. Additional details about datasets and compilation criteria are provided in Appendix~\ref{app:datasets}.

\paragraph{Baselines.}
The proposed SIC method is compared against 10 different strategies, spanning linear, neural net, tree-based, and PFN approaches. CoxPH~\citep{cox72} uses a linear model with a PH assumption; DeepSurv~\citep{katzman18} applies a neural network to encode the non-linear dependencies under PH assumption; DeepHit~\citep{lee18} models the survival distribution by discretizing the output times into a fixed number of segments using neural networks; Random Survival Forests (RSF) and XGBoost (XGB)~\citep{chen16} are tree-based methods adapted to survival analysis; SuMo-Net~\citep{Rindt21} internalizes the discretization within a hazard network; TabICL + Stacking transforms the survival analysis into a binary classification problem using discrete-hazard Survival Stacking~\citep{craig21} and applies classification PFN TabICL~\citep{qu25}. Similarly, the concurrent work of \citet{kim26r} evaluates classification models under both discrete-hazard (DH) and cumulative-failure (CF) formulations, including MITRA~\citep{mitra}, TabPFNv2.5~\citep{grinsztajn25}, and XGBoost~\citep{xgboost}. We report MITRA\_CF, the best-performing configuration in their experiments. Finally, the concurrent SurvPFN~\citep{boehm26} and SurvivalPFN~\citep{qi26} introduce a survival prior PFN, similar to ours.

\paragraph{Evaluation.}
We employ nested 5-fold cross-validation, with 3-fold cross-validation for hyperparameter tuning. We assess both discrimination and calibration performance using the integrated Brier score (IBS); discrimination performance using the time-dependent C-index (td-CI) and marginal calibration using D-calibration (D-Cal)~\citep{haider20} by counting the number of misscalibrated folds ($p < 0.05$) across all datasets, as in~\citet{lee24}. Additional information regarding td-CI, IBS, and D-Cal is provided in Appendix ~\ref{app:metrics}. 

For hyperparameter optimization of the baseline methods, we use the Tree-Structured Parzen Estimator (Bayesian optimization) in Optuna~\citep{optuna} with 50 trials or 5h per fold for all models, setting td-CI as the optimization target, for comparing the td-CI, and IBS for comparing the IBS performance~\footnote{\citet{chen26} showed that the choice of validation criterion has a substantial impact, since selecting models based on IBS generally yields lower td-CI for tunable models and vice versa. Following their evaluation protocol, we compare our untuned method against the strongest tuned baseline achievable under each model-selection strategy.} The hyperparameter search space is reported in Appendix~\ref{app:hyperparams}. The PFN models, including SIC, do not require \textit{any} hyperparameter tuning and are evaluated using a single forward pass. 

To assess baselines across all datasets, we employ an Elo rating system~\citep{elo}, similar to TabArena~\citep{erickson25}. Elo is a pairwise-comparison-based rating system in which the model's rating predicts its expected win probability against others, and where only relative differences between Elo scores are meaningful, as they predict win-rates between pairs of methods. We pin Elo 1000 to CoxPH (default) and perform 200 rounds of dataset bootstrapping to obtain 95\% confidence intervals. We retain CoxPH (default) as the conventional unregularized Cox baseline even though it frequently fails to converge on the datasets. When a method fails to produce a finite score on a dataset, that outcome is counted as a loss against any opponent with a finite score (see further details about failures and convergence in Appendix~\ref{app:failures}).

\subsection{Results}
Figure~\ref{fig:results} presents the performance of SIC relative to the baselines in terms of IBS and td-CI. The additional D-cal results are provided in Appendix~\ref{app:calibration}. The numerical values are provided in Appendix~\ref{app:allresults}.

\paragraph{Survival Analysis requires a separate PFN formulation.}
Naively applying off-the-shelf PFN with Stacking (TabICL + Stacking) exhibits substantially worse performance. Stacking~\citep{craig21} transforms survival data by converting the task into binary classification. However, transformation greatly enlarges the dataset, making it difficult to fit within the fixed context limits of classification PFNs, such as recommended 100,000 rows for TabICLv1 (e.g., \texttt{METABRIC}, originally 1904 rows, expands to 1,131,884 rows after stacking). As a result, large support sets must be subsampled, and test sets often require multiple inference batches, increasing cost and reducing performance. This justifies the necessity of changing the stacking formulation or employing a dedicated survival prior. Concurrent work, MITRA\_CF~\citep{kim26r}, demonstrates that the limitations of naive stacking for PFN can also be mitigated by reformulating the stacking procedure itself. This approach achieves Elo rating in td-CI comparable to SIC, although it remains worse in IBS and D-cal (see Appendix~\ref{app:calibration}). This improvement comes at a higher computational cost: MITRA\_CF requires approximately $10\times$ longer inference due to the expanded stacking representation. SurvPFN and SurvivalPFN avoid this expansion by incorporating a survival prior directly, as does SIC, and are consequently substantially faster than MITRA\_CF. SurvivalPFN remains slightly slower than SIC, which we attribute to its TabDPT encoder being slower than the TabICL one used by SIC. SurvPFN is slightly faster than SIC due to the subsampling of the context window (see details in Appendix~\ref{app:failures}). In terms of predictive performance, SurvPFN and SurvivalPFN perform slightly worse than SIC in IBS and D-cal, with a larger gap in td-CI.We hypothesize that this gap is partly due to the discretized output head of SurvivalPFN and SurvPFN, while SIC internalizes the discretization..

\paragraph{SIC achieves the strongest performance.} SIC achieves the highest Elo rating in IBS and td-CI, while being the fastest by inputting the entire context. The D-calibration results further show that this predictive performance does not come at the expense of calibration: SIC has, on average, 0.7 miscalibrated folds, close to the best-performing RSF (0.4) and DeepSurv (0.6). Additional aggregate metrics, including mean and median rank, win rate, and improvability (see Appendix~\ref{app:allresults}) confirm the same ordering, with SIC first on every metric for IBS and td-CI, suggesting consistently strong performance across survival tasks using only a single model.

\paragraph{SIC is effective on small- and medium-sized datasets.} To analyze performance across datasets with respect to dataset size, we visualize per-dataset ranks for each method and summarize them in Appendix~\ref{app:rank} for IBS and td-CI. SIC tends to attend the higher ranks (1-5) for small and medium-sized datasets (darker colors), while attending lower ranks (6-10) on larger datasets (lighter colors). SurvPFN, SurvivalPFN, and MITRA\_CF exhibit similar trends, suggesting that PFN-based approaches are particularly competitive in the small- and medium-data regime. Further analysis confirms observations reported in the literature: nonlinear methods, such as DeepHit, typically perform better with more data, whereas simpler models, such as CoxPH, are more effective on small datasets. Tree-based methods, such as RSF, remain strong defaults across diverse datasets.

\paragraph{SIC is fast.}
PFNs completely avoid parameter optimization, making predictions in a single forward pass using training data as context. This reveals two strengths. First, even though PFNs can fit highly non-linear functional forms, they do not require large datasets, making them particularly interesting for tasks with limited data. Second, the PFN is known to offer a huge speedup for non-survival tasks ~\citep{hollmann25, qu25}. We confirm that these efficiency gains also carry over to SIC in survival analysis. SIC requires a total of 46.8m for all 58 datasets, while tree-based RSF, linear CoxPH, and neural SuMo-Net require 62.1h, 22.2h, and 143h, respectively, to achieve closer performance. This implies a substantial speed-up, making it a promising candidate for settings where fast, reliable, and well-tuned predictions are crucial, e.g., continually refreshed pipelines in clinical operations~\citep{susetzky26} or high-throughput hypothesis testing of many endpoints in biobanks~\citep{hager26}.

\paragraph{SIC is not restricted to a particular covariate or survival distribution.}
We observe strong performance of the simple CoxPH model, suggesting that structural assumptions can be particularly beneficial for small datasets. However, this advantage can disappear when these assumptions do not hold. To show this, we simulate six survival datasets: (Exp-PH-100/1000/10000) with Exponential-PH assumptions for survival times with 100/1000/10000 samples; (Lognormal-AFT-DAG-100/1000/10000) simple non-linear covariate dependence for Lognormal-AFT survival times generation (details in Appendix~\ref{app:simulation}). Table~\ref{tab:simulation} reports IBS and td-CI for SIC and representative models from different model classes: linear CoxPH, tree-based RSF, and neural SuMo-Net. In the low-sample setting where the PH assumption holds (Exp-PH-100), CoxPH is particularly advantageous over the more flexible baselines. With increasing sample size, however, neural and tree-based methods become competitive with CoxPH. In the non-Exp-PH settings, where linear covariate dependencies and PH survival distribution break, CoxPH performance deteriorates substantially. Neural and tree-based methods perform better due to their flexible formulations, but require more samples for stable performance. All baselines require hyperparameter tuning and, thus, more time for fitting. SIC bridges the investigated regimes by delivering fast and stable performance across all simulation experiments.

\begin{table}[t]
  \caption{Integrated Brier Score (IBS) and time-dependent C-index (td-CI) for the proposed Survival In-Context (SIC), CoxPH~\citep{cox72}, RSF~\citep{ishawaran08}, and SuMo-Net~\citep{Rindt21} on simulated datasets: (Exp-PH-100/1000/10000), with exponentially distributed survival times under proportional hazards (PH) assumptions with 100/1000/10000 samples; (Lognormal-AFT-DAG-100/1000/10000), with lognormal survival times and non-linear covariate dependence. Although CoxPH performs strongly on real-world datasets, its performance degrades in the presence of nonlinear covariate effects and violations of the PH assumption. SIC consistently delivers fast and stable performance across different covariate and survival time distributions. Best in \textbf{bold}, second best in \textit{italics}.}
  \label{tab:simulation}
  \centering
  \scriptsize
  \setlength{\tabcolsep}{3pt}
  \begin{tabular}{lcccccccc}
    \toprule
    & \multicolumn{4}{c}{IBS ($\downarrow$)} & \multicolumn{4}{c}{td-CI ($\uparrow$)} \\
    \cmidrule(lr){2-5} \cmidrule(lr){6-9}
    Dataset & SIC & CoxPH & RSF & SuMo-Net & SIC & CoxPH & RSF & SuMo-Net \\
    \midrule
    Exp-PH-100 & \textit{0.19{\tiny$\pm$0.03}} & \textbf{0.17{\tiny$\pm$0.03}} & \textit{0.19{\tiny$\pm$0.03}} & 0.21{\tiny$\pm$0.04} & \textit{0.64{\tiny$\pm$0.12}} & \textbf{0.66{\tiny$\pm$0.05}} & 0.61{\tiny$\pm$0.05} & \textit{0.64{\tiny$\pm$0.11}} \\
    Exp-PH-1000 & \textbf{0.16{\tiny$\pm$0.01}} & \textbf{0.16{\tiny$\pm$0.01}} & \textit{0.17{\tiny$\pm$0.00}} & \textbf{0.16{\tiny$\pm$0.01}} & \textbf{0.73{\tiny$\pm$0.01}} & \textbf{0.73{\tiny$\pm$0.01}} & \textit{0.70{\tiny$\pm$0.03}} & \textbf{0.73{\tiny$\pm$0.02}} \\
    Exp-PH-10000 & \textbf{0.16{\tiny$\pm$0.00}} & \textbf{0.16{\tiny$\pm$0.00}} & \textbf{0.16{\tiny$\pm$0.00}} & \textbf{0.16{\tiny$\pm$0.00}} & \textbf{0.74{\tiny$\pm$0.00}} & \textbf{0.74{\tiny$\pm$0.01}} & \textit{0.73{\tiny$\pm$0.01}} & \textbf{0.74{\tiny$\pm$0.01}} \\
    Lognormal-AFT-DAG-100 & \textbf{0.19{\tiny$\pm$0.01}} & 0.25{\tiny$\pm$0.04} & \textit{0.22{\tiny$\pm$0.06}} & 0.26{\tiny$\pm$0.06} & \textbf{0.71{\tiny$\pm$0.05}} & 0.64{\tiny$\pm$0.13} & 0.67{\tiny$\pm$0.07} & \textit{0.68{\tiny$\pm$0.11}} \\
    Lognormal-AFT-DAG-1000 & \textbf{0.14{\tiny$\pm$0.01}} & \textit{0.20{\tiny$\pm$0.00}} & \textbf{0.14{\tiny$\pm$0.01}} & \textbf{0.14{\tiny$\pm$0.01}} & \textbf{0.79{\tiny$\pm$0.02}} & 0.65{\tiny$\pm$0.01} & \textit{0.78{\tiny$\pm$0.02}} & \textbf{0.79{\tiny$\pm$0.02}} \\
    Lognormal-AFT-DAG-10000 & \textbf{0.14{\tiny$\pm$0.00}} & \textit{0.20{\tiny$\pm$0.00}} & \textbf{0.14{\tiny$\pm$0.00}} & \textbf{0.14{\tiny$\pm$0.00}} & \textbf{0.79{\tiny$\pm$0.01}} & 0.67{\tiny$\pm$0.01} & \textit{0.78{\tiny$\pm$0.01}} & \textbf{0.79{\tiny$\pm$0.01}} \\
    \midrule
    Total time (h) & \textbf{0.15} & \textit{0.91} & 54.32 & 15.77 & \textbf{0.15} & \textit{0.92} & 53.91 & 17.56 \\
    \bottomrule
  \end{tabular}
\end{table}

\subsection{Discussion and Ablations}
\label{res:ablation}

\paragraph{Priors.} Appendix~\ref{app:ablation:prior} presents an ablation of continuous-only, discrete-only, and mixed priors. While results vary across datasets and no single prior uniformly dominates, the mixed prior has better performance on average. As discussed in Section~\ref{sec:background}, PFNs critically depend on the choice of prior. A well-specified prior enables accurate approximation of the PPD, whereas an overly simple prior limits expressivity and an overly broad or complex prior can waste Monte Carlo training effort on unrealistic regions of the function space. Our mixed prior is designed to balance these considerations by combining complementary inductive biases rather than committing to a single prior family. 

\paragraph{Curriculum Learning.}
We evaluate the curriculum learning strategy proposed in~\cite {qu25} by comparing two stages in Appendix~\ref{app:ablation:cl}. Stage 2 slightly improves the results on the larger datasets, but the improvement is more modest in our benchmark than those reported for classification~\citep{qu25}. 
Since this could be due to the generally lower sample sizes in our evaluation datasets, and to ensure robustness on future unseen datasets, we used the model obtained after stage 2 for all subsequent evaluations. 

\paragraph{Survival head.} The SIC framework is not limited to a specific survival head, allowing for flexible architectural choices. We evaluate DeepEH~\citep{zhong21} and DeepHit~\citep{katzman18} as alternative survival heads. As shown in Appendix~\ref{app:ablation:heads}, the alternative survival heads exhibit comparable or inferior performance across datasets compared to SuMo++. We therefore adopt SuMo++ due to its continuous-time flexibility and non-parametric formulation.

\paragraph{Informative censoring and identifiability.}
Our prior assumes conditionally independent, non-informative censoring by construction. In real-world settings, however, event and censoring risks may share unobserved causes (e.g., unobserved treatment intolerance increases both the risk of event and dropping out of follow-up), violating conditional independence and inducing informative censoring. In this case, the right-censored likelihood in Equation~\ref{eq:nll} is misspecified for the true event-time distribution~\citep{Rindt21} and the true event survival curve is generally not identifiable from the observed data~\citep{tsiatis75}. All tuned baselines optimize the same likelihood and hence hold the same assumptions. Training on the true event times, which are available during synthetic prior generation, would reduce the objective to the uncensored log-likelihood (the event term in Equation~\ref{eq:nll}), providing well-specified supervision for the true event-time distribution. However, the survival function would remain non-identifiable from the observed context alone. How an in-context learner behaves in this setting remains an open question. Our survival prior generation and method can be extended to informative censoring and true-event-time supervision, as explored in preliminary experiments in Appendix~\ref{app:identify}. A systematic theoretical and empirical investigation of PFN with conditionally dependent censoring remains for future work. Further limitations and outlook are discussed in Appendix~\ref{app:limitations}.

\section{Conclusion}
\label{sec:conclusion}
We introduce a Survival In-context Model (SIC), the prior-fitted in-context learning model tailored to survival analysis. Our approach is enabled by a novel method for synthesizing survival data using structural causal models (SCMs)~\citep{pearl09}, the extended hazard assumption~\citep{amoli87}, and monotonic networks~\citep{Rindt21}. Together, these components allow the generation of covariates and time-to-event outcomes without relying on real-world datasets, while ensuring statistical properties. We extensively evaluate SIC on multiple real-world datasets and compare it against established baselines. Our single SIC model outperforms or achieves comparable performance to dataset-specific classical, deep learning, and concurrent PFN baselines while requiring no hyperparameter tuning and being orders of magnitude faster. This substantially simplifies deployment by eliminating the need to build a dedicated machine learning pipeline, reducing training overhead, and enabling prediction with a single inference pass.

\bibliography{main}
\bibliographystyle{tmlr}


\clearpage
\appendix
\etocdepthtag.toc{appendix}

\begingroup

  \etocsettagdepth{main}{none}
  \etocsettagdepth{appendix}{section}
  \etocsetnexttocdepth{section}

  \etocsettocstyle{\section*{Appendix Contents}}{}
  \tableofcontents
\endgroup

\clearpage

\section{Additional Related Work}
\label{app:relatedwork}

\paragraph{Synthetic Data Generation for Survival Analysis.}
Early work on survival data generation focuses on the statistical parametrization of survival times~\citep{bender05, austin12}, assuming simple stochastic processes to generate the covariates. Subsequent work introduces more expressive latent-variable models, such as deep exponential families~\citep{ranganath16}, later extended to broader distributional families~\citep{miscouridou18}, while still focusing primarily on survival times. SurvivalGAN~\citep{norcliffe23} and the Ashhad framework~\citep{ashhad25} shift attention to generating covariates using conditional generative models. However, these approaches require conditioning on existing real-world datasets, which are scarce in the survival analysis domain. Our approach focuses on both generating covariates without relying on underlying real-world data and on modeling time-to-event outcomes by incorporating statistical properties and prior knowledge of common distributions encountered in survival analysis.

\section{Details on Continuous Survival Prior}
\label{app:prior:continuous}
\subsection{Time Derivation under Extended Hazard Assumption}
\label{app:T}
Under the extended hazard assumption, $h(t \mid x)=h_0\left(t e^{\eta_1}\right) e^{\eta_2}$~\citep{amoli87}, we derive $T$: 

\begin{equation}
            H(t \mid x)=\int_0^t e^{\eta_2}\, h_0(\underbrace{u\eta_1}_{=:v})\, d u \\
\end{equation}
\begin{equation}
            H(t \mid x)=\int_{0}^{t e^{\eta_1}} e^{\eta_2}\, h_0(u)\, e^{-\eta_1}\, d u  \text{, where } \\
            \begin{cases}
            v = u e^{\eta_1} \\
            u = v e^{-\eta_1} \\
            d u = e^{-\eta_1}\, d v
            \end{cases}
            \qquad
            \begin{cases}
            u=0 \;\Rightarrow\; v=0 \\
            u=t \;\Rightarrow\; v=t e^{\eta_1}
            \end{cases} \\
\end{equation}
\begin{equation}
            H(t \mid x)=e^{\eta_2-\eta_1}\int_0^{t e^{\eta_1}} h_0(v)\, d v =e^{\eta_2-\eta_1}\, H_0\!\left(t e^{\eta_1}\right)
\end{equation}

\begin{equation}
        S(t \mid x)=e^{-H(t \mid x)} \Rightarrow 
        \log \underbrace{S(t \mid x)}_{=:U}=-e^{\eta_2-\eta_1}\, H_0\!\left(T e^{\eta_1}\right) \Rightarrow
        H_0\!\left(T e^{\eta_1}\right)=e^{\eta_1-\eta_2}(-\log U) \\
\end{equation}

\begin{equation}
    T=e^{-\eta_1}\, H_0^{-1}\!\left(e^{\eta_1-\eta_2}(-\log U)\right)
\end{equation}

\subsection{Inverse Baseline Cumulative Hazard Distributions}
\label{app:H0-1}
Table~\ref{tab:invhazard} provides the inverse cumulative hazard $H_0^{-1}$ for baseline survival distributions used in Continuous Prior.
\begin{table}[H]
\caption{Inverse cumulative hazard $H_0^{-1}(y)$ for baseline survival distributions.}
\label{tab:invhazard}
\begin{center}
\begin{small}
\begin{sc}
\begin{tabular}{ll}
\toprule
Distribution & $H_0^{-1}(y)$ \\
\midrule
Weibull
& $\displaystyle \alpha\, y^{1/\beta} \text{; } \alpha\ > 0 \text{, } \beta > 0$ \\[6pt]
Gompertz
& $\displaystyle \frac{1}{\alpha}\,
  \ln\!\left(1 + \frac{\alpha}{\beta}\,y\right) \text{; } \alpha\ > 0 \text{, } \beta \neq 0$ \\[6pt]
Lognormal
& $\displaystyle \exp\!\Big(
    \alpha
    + \beta\,\Phi^{-1}(1 - e^{-y})
  \Big) \text{; } \alpha\ > 0$ \\[6pt]
Log-logistic
& $\displaystyle \alpha\,\big(e^{y} - 1\big)^{1/\beta} \text{; } \alpha\ > 0 \text{, } \beta > 0$ \\[6pt]
Birnbaum--Saunders
& $\displaystyle \alpha\left[
    \tfrac{1}{2}\Big(
      \beta\,\Phi^{-1}(1 - e^{-y})
      + \sqrt{(\beta\,\Phi^{-1}(1 - e^{-y}))^{2} + 4}
    \Big)
  \right]^2 \text{; } \alpha\ > 0 \text{, } \beta > 0$ \\
\bottomrule
\end{tabular}
\end{sc}
\end{small}
\end{center}
  \vskip -0.1in
\end{table}

\subsection{Control Parameters.} 
To avoid different scaling of the distribution curve, we apply scale anchoring. For each dataset, we enforce that each baseline survival curve passes the same randomly selected point $(t^*, q^*)$, i.e., $S_0(t^*)=q^*$. This means we define each scale parameter using anchoring and the shape parameter is sampled for each survival curve independently. For example, for Weibull distribution, $H_0=\frac{t}{\alpha}^\beta$, and the inverse $H_0^{-1}(t) = \alpha t^{1/\beta}$. This means:
\begin{equation}
    S_0(t) = exp(-(\frac{t}{\alpha})^\beta) = q^* \implies
    -(\frac{t^*}{\alpha})^\beta ) = log(q^*) \implies \alpha = \frac{t^*}{(-log(q^*))^{1/\beta}},\\
\end{equation}
where $\beta > 0$ is a shape parameter and sampled independently. For the other distributions, we perform the same procedure. 

We set the control parameters for the final generation as follows: 
$q^* \sim \mathcal{U}(0,1)$, $t^* \sim \mathcal{U}(0,1000)$, Weibull $\beta \sim \mathcal{U}[0.5, 5.0]$, Gompertz $\beta \sim \mathcal{U}[0.0, 0.3]$, Lognormal $\beta \sim \mathcal{U}[0.3, 1.5]$, Log-Logistic $\beta \sim \mathcal{U}[0.5, 5.0]$, Birnbaum-Saunders $\beta \sim \mathcal{U}[0.5, 5.0]$. All distributions are sampled equally random. The administrative censoring quantile is sampled from  $\mathcal{U}[0.1, 0.9]$. Note that these parameter ranges are selected to ensure stable numerical behavior, but may be modified depending on the application.

\section{Post-hoc Analysis of Censoring Mechanisms}
\label{app:posthoccheck}

We verify on generated data that (i) marginally independent censoring does not depend on $X$,
(ii) feature-dependent censoring depends on $X$, and (iii) both are non-informative,
$C \perp T \mid X$.

\paragraph{Setup.}
For each prior (continuous and discrete), we sample one batch of $512$ datasets ($N=1024$ subjects each)
per censoring configuration and group them by how censoring was generated: \emph{independent}
($\eta^{C}=0$), \emph{feature-dependent} ($\eta^{C}$ from the SCM on $X$) and, as a reference, \emph{non-restricted} (the $\eta^{C}$ and $\eta$ may have a shared confounder that is not in X, leading to informative censoring with $C \not\perp T \mid X$).
 
We work with log times $Z = \log t$, on which the multiplicative effect of the risk scores becomes approximately additive, and fit per dataset three OLS regressions. To test dependence on $X$, we apply overall F-tests for all $p$ coefficients of the regression equation. (a) We test the dependence $T \perp X$ using $Z_T = a + X\beta + \varepsilon$ ($H_0\!: \beta = 0$). Further, we test (b) $C \perp X$ using $Z_C = a + X\beta + \varepsilon$ using the same null hypothesis $H_0\!: \beta = 0$.  To investigate conditional independence ($C \perp X$) (c), we use as regression equation $Z_T = a + X\beta + \gamma Z_C + \varepsilon$ with null hypothesis $H_0: \gamma=0$ where we apply t-tests for targeted testing. We report the share of datasets in which
$H_0$ is rejected at $\alpha=0.05$ with Wilson 95\% confidence intervals. Since (c) adjusts for $X$ only linearly, the non-linear effects of $X$ shared by $T$ and $C$ can affect it. The rate for feature-dependent censoring, where $C \perp T \mid X$ holds by construction, may serve as its empirical reference point.

\paragraph{Results.}
In both priors, independent censoring rejects (b) and (c) with rate 3\% to 8\%, consistent with $C \perp (T, X)$. Feature-dependent censoring rejects (b) in 25\% (continuous) and 29\% (discrete) of datasets, while (c) stays at the 5\% and 8\%, respectively. The censoring depends on $X$ but remains non-informative. The non-restricted prior (allows the shared unobserved confounders, and, thus, is not used in SIC training) rejects (c) in 15\% (continuous) and 23\% (discrete) of datasets versus 5\% and 8\% for feature-dependent censoring, showing that the test detects residual dependence between $T$ and $C$. These results provide an additional empirical sanity check for consistency with the intended dependence construction in the prior. As the tests capture only linear associations, they should not be interpreted as a complete formal verification of (conditional) independence.

\begin{table}[H]
\centering
\small
\caption{Ablation study of censoring mechanisms. Share of datasets (percentage, 95\% CI) rejecting $H_0$ at $\alpha=0.05$.  The proposed Survival In-context (SIC) uses independent and feature-dependent mechanisms. }
\label{tab:censoring_check}
\begin{tabular}{lccccc}
\toprule
Regime & $C\!\perp\! X$ & $C\!\perp\! T\!\mid\! X$
  & (a) $T\sim X$ & (b) $C\sim X$ & (c) $C\not\perp T \mid X$ \\
\midrule
\multicolumn{6}{l}{\textit{Continuous prior}} \\
Independent            & \cmark & \cmark & 47 [41, 54] & 3 [2, 7]    & 6 [4, 10] \\
Feature-dependent      & \xmark & \cmark & 47 [42, 52] & 25 [21, 29] & 5 [3, 7] \\
Non-restricted (not used in SIC) & \xmark & \xmark & 43 [38, 48] & 35 [31, 40] & 15 [12, 19] \\
\midrule
\multicolumn{6}{l}{\textit{Discrete prior}} \\
Independent            & \cmark & \cmark & 50 [44, 57] & 8 [5, 12]   & 4 [2, 8] \\
Feature-dependent      & \xmark & \cmark & 49 [43, 54] & 29 [25, 34] & 8 [6, 11] \\
Non-restricted (not used in SIC) & \xmark & \xmark & 46 [41, 51] & 41 [36, 46] & 23 [19, 28] \\
\bottomrule
\end{tabular}
\end{table}

\section{Details on Discrete Survival Prior}
\label{app:surv_prior}
\subsection{Anchoring Endpoints of Survival Curve.}
\label{app:surv_prior:anchor}
To obtain a valid survival curve on the full normalized time range, we anchor the endpoints. In particular, we normalize the network output such that the survival curve satisfies $S(t=0, \eta) = 1$ and $S(t=1|\eta) = 0$:
\[
    S(t|\eta) = \frac{f_\omega(\eta, 1) - f_\omega(\eta, t)}{f_\omega(\eta, 1) - f_\omega(\eta, 0)}
\]

\subsection{Weight Initialization.} 
\label{app:surv_prior:weightinit}
We use the Gamma distribution $\Gamma(k, \theta)$ to initialize positive weights. $k>0$ is a shape and $\theta>0$ is a scale parameter. Note that in the Jeanselme et al. implementation~\footnote{\url{https://github.com/Jeanselme/SuMo-net}} of SuMo-Net~\citep{Rindt21}, the weights are initialized with $\mathcal{N}(0, \sigma^2)$ and then squared, which corresponds to chi-squared or a special case of $\Gamma(k=0.5, \theta=2\sigma^2)$ distribution.

Standard Xavier initialization selects the weight scale relative to the number of input neurons (fan-in) so that the pre-activation variance remains approximately stable across layers. Since our weights are positive and therefore not zero-centered, we use a Xavier fan-in correction on the second moment $\mathbb{E}[W^2]$, rather than on $\mathrm{Var}(W)$, to control the pre-activation variance through the gain $\alpha$ and fan-in $n$. The variance of the pre-activation $z_i$ in the layer $l$, assuming independent weights and approximately centered activations $\mathbb{E}[h_j^{(\ell-1)}]\approx 0$ (see the first layer initialization paragraph in Section ~\ref{sec:method}):

\begin{align*}
\mathrm{Var}(z_i^l)
&= \mathrm{Var}\left(\sum_{j=1}^n w_{ij}^l h_j^{l-1}\right) = \sum_{j=1}^n \mathrm{Var}(w_{ij}^l h_j^{l-1}) \\
&= \sum_{j=1}^n \Big(
    \mathbb{E}[w_{ij}^l]^2 \mathrm{Var}(h_j^{l-1})
    + \mathbb{E}[h_j^{l-1}]^2 \mathrm{Var}(w_{ij}^l) + Var(w_{ij}^l)Var(h_j^{l-1})\Big) \\
&= \sum_{j=1}^n \mathrm{Var}(h_j^{l-1}) ( \mathbb{E}[w_{ij}^l]^2 + Var(w_{ij^l}) = \sum_{j=1}^n \mathrm{Var}(h_j^{l-1}) ( \mathbb{E}[(w_{ij}^{l})^2] = \\
&= n\mathrm{Var}(h_j^{l-1}) ( \mathbb{E}[(w_{ij}^{l})^2]
\end{align*}
We represent $\mathbb{E}[w_{ij}^2] = E[W^2] = \alpha^2/n$. Using $W \sim \Gamma(k,\theta)$ and the moments $\mathbb{E}[W] = k\theta$ and $Var(W) = k\theta^2$, we derive the scale $\theta$: 
\[
    \mathbb{E}[W^2] = \mathbb{E}[W]^2 + Var(W) = k(k+1)\theta^2 \implies \theta = \frac{\alpha}{\sqrt{n k(k+1)}}
\]

\subsection{Censoring and observed outcomes.}
\label{app:surv_prior:censor}
Censoring times are generated using a second positive monotone network $g_\psi(t)$ that depends only on time. Its normalized survival curve $S_\psi(t)$ is constructed analogously to $S_\omega(t \mid \eta)$, and
censoring times $C_i$ are sampled by the same exponential interpolation procedure. To generate datasets on heterogeneous time scales, normalized event and censoring times are multiplied by a dataset-specific scale $t_{\max} \sim \mathrm{LogUniform}(1, 10^4)$

Finally, we apply administrative censoring (end of study due to limited follow-up time or a limited number of events). As in the continuous prior, we compute the cutoff time $T_{adminType1}$ at the random quantile of the time or $T_{adminType2}$ at the random quantile of events and truncate observation times at these cutoffs, resulting in $T'_{C} = min(C, T_{adminType1}, T_{adminType2})$. The final observed times are $T_{obs} = min(T,T'_{C})$ with $e=\mathbbm{1}\{T<T'_{C}\}$.

\subsection{Control Parameters.} 
\label{app:surv_prior:params}
\begin{figure}[H]
\centering

\includegraphics[height=0.9\textheight]{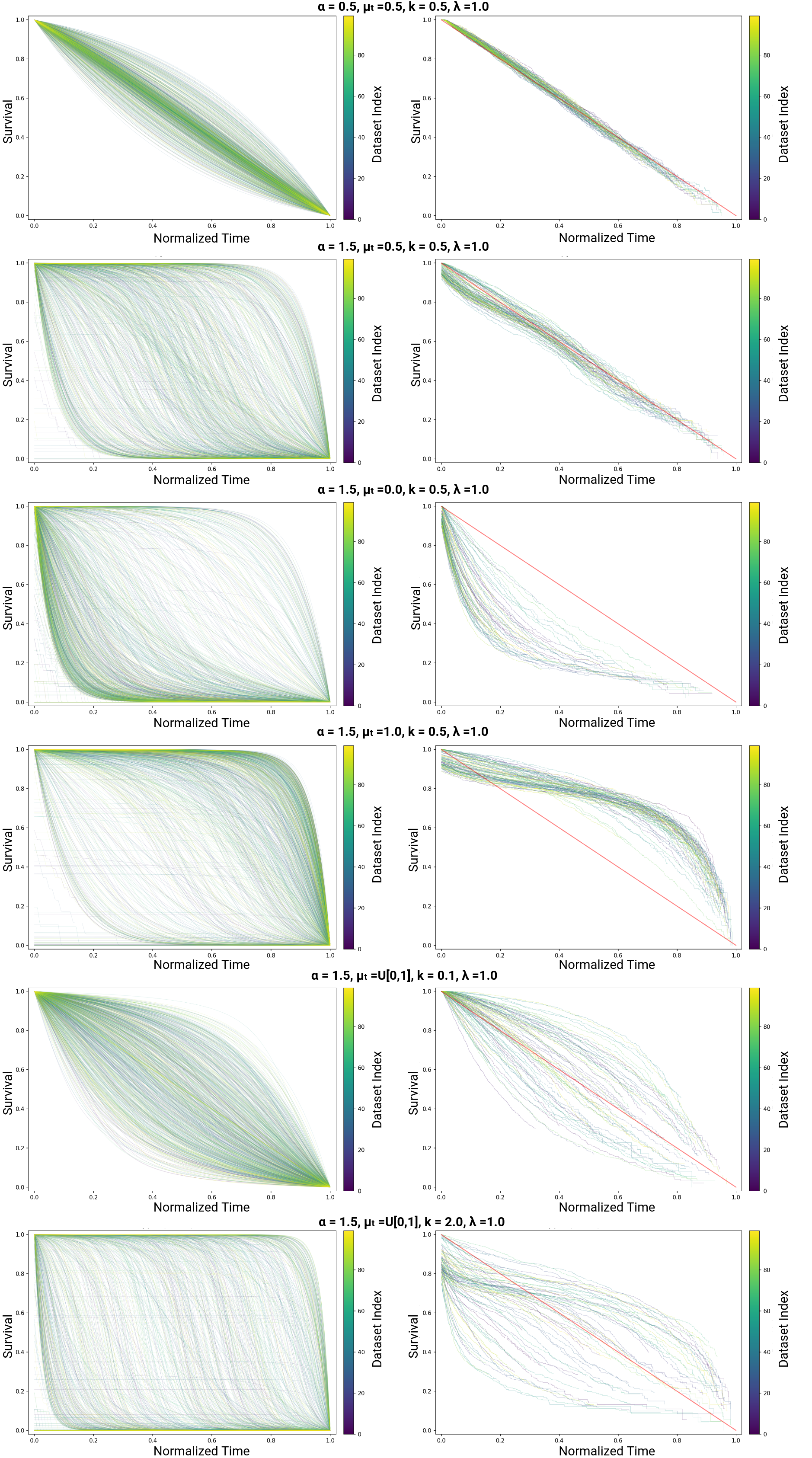}
\caption{Visualization of control parameters for a discrete survival prior: survival (left) and Kaplan-Meier (right) curves for a batch of 100 datasets.}
\label{fig:mono_ablation}
\end{figure}

Together, $\alpha$, $k$, $\mu_t$, and $\lambda$ define a prior over survival curve shapes. In particular, $\alpha$ controls global steepness, $k$ controls
weight concentration and tail-heaviness, $\mu_t$ controls the time-location of
nonlinear transitions, and $\lambda$ controls covariate-driven separation
between survival curves. The examples of the concrete parameter values and how they control the survival curves are provided in Figure~\ref{fig:mono_ablation}. 

We set the control parameters for the final generation as follows: $\alpha \sim \mathcal{U}[0.5, 1.5]$, $k \sim \mathcal{U}[0.1, 2.0]$, $\mu_t \sim \mathcal{U}[0.0, 1.0]$, $\lambda \sim \mathcal{U}[0.1, 1.0]$. Note that these parameter ranges are selected to ensure stable numerical behavior, but may be modified depending on the application.

\section{Training Details}
\label{app:training}
We train the model in 2 stages. In stage 1, we use fixed 1,024 samples per table over 10K steps to learn a first good general representation.  In stage 2, we use a variable sample size $\in[10, 10000]$ over 2K steps to further account for the variable size of considered datasets. 

Each step consists of 512 artificially generated datasets with  $p \leq 100$ covariates. Automatic mixed precision is applied globally. During training, the encoder is initialized using the original TabICL weights trained for classification. We use AdamW~\citep{loshchilov17} optimizer with cosine decay learning rate peaking at $10^{-4}$ for stage 1, polynomial decay from $10^{-5}$ to $10^{-6}$ for stage 2. The pretraining took 120 h for stage 1 and 55 h for stage 2 on one A100 80GB GPU. 

\begin{figure}[htbp]
    \centering
    \begin{subfigure}[b]{0.32\textwidth}
        \centering
        \includegraphics[width=\textwidth]{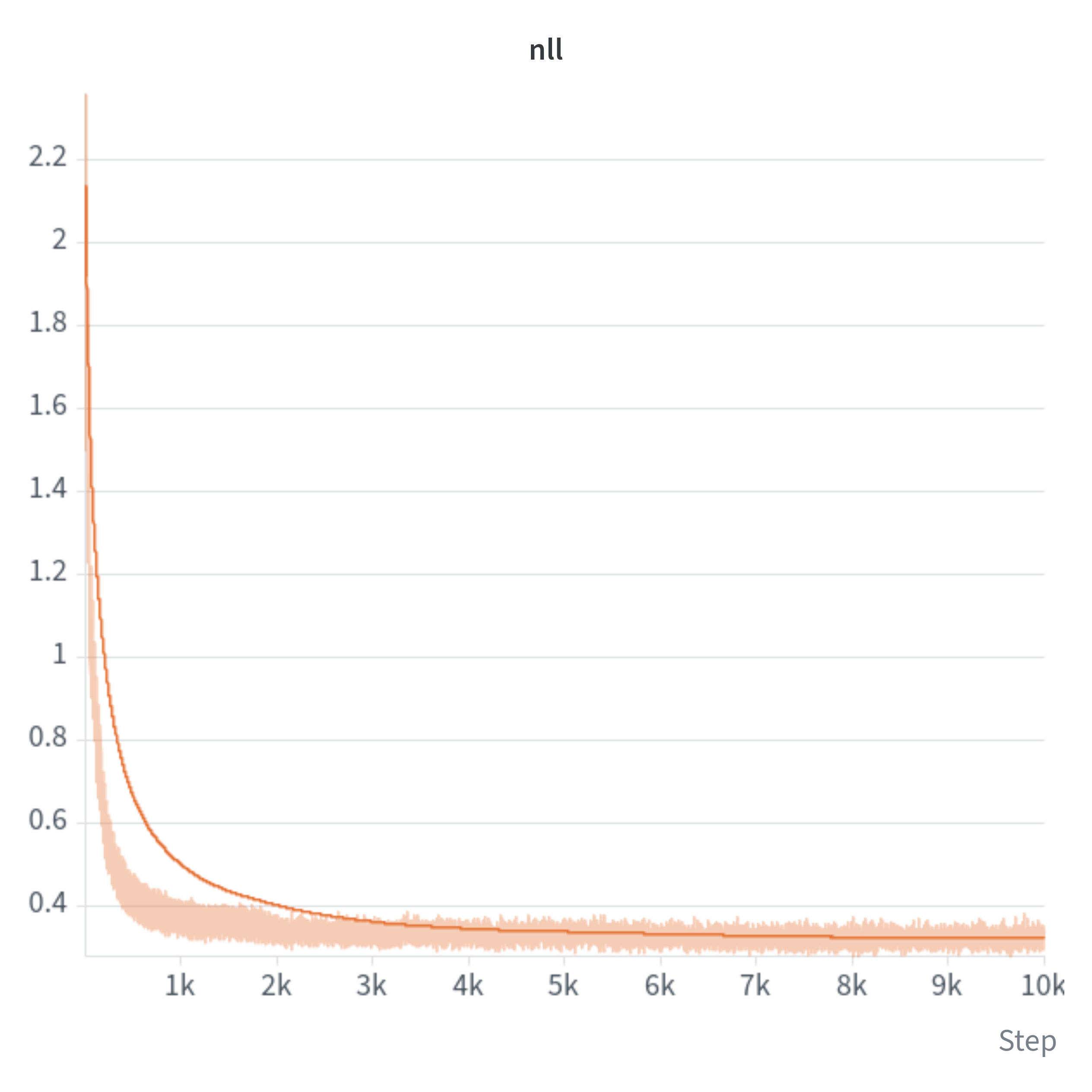}
        \caption{Training Loss.}
        \label{fig:image1}
    \end{subfigure}
    \hfill
    \begin{subfigure}[b]{0.32\textwidth}
        \centering
        \includegraphics[width=\textwidth]{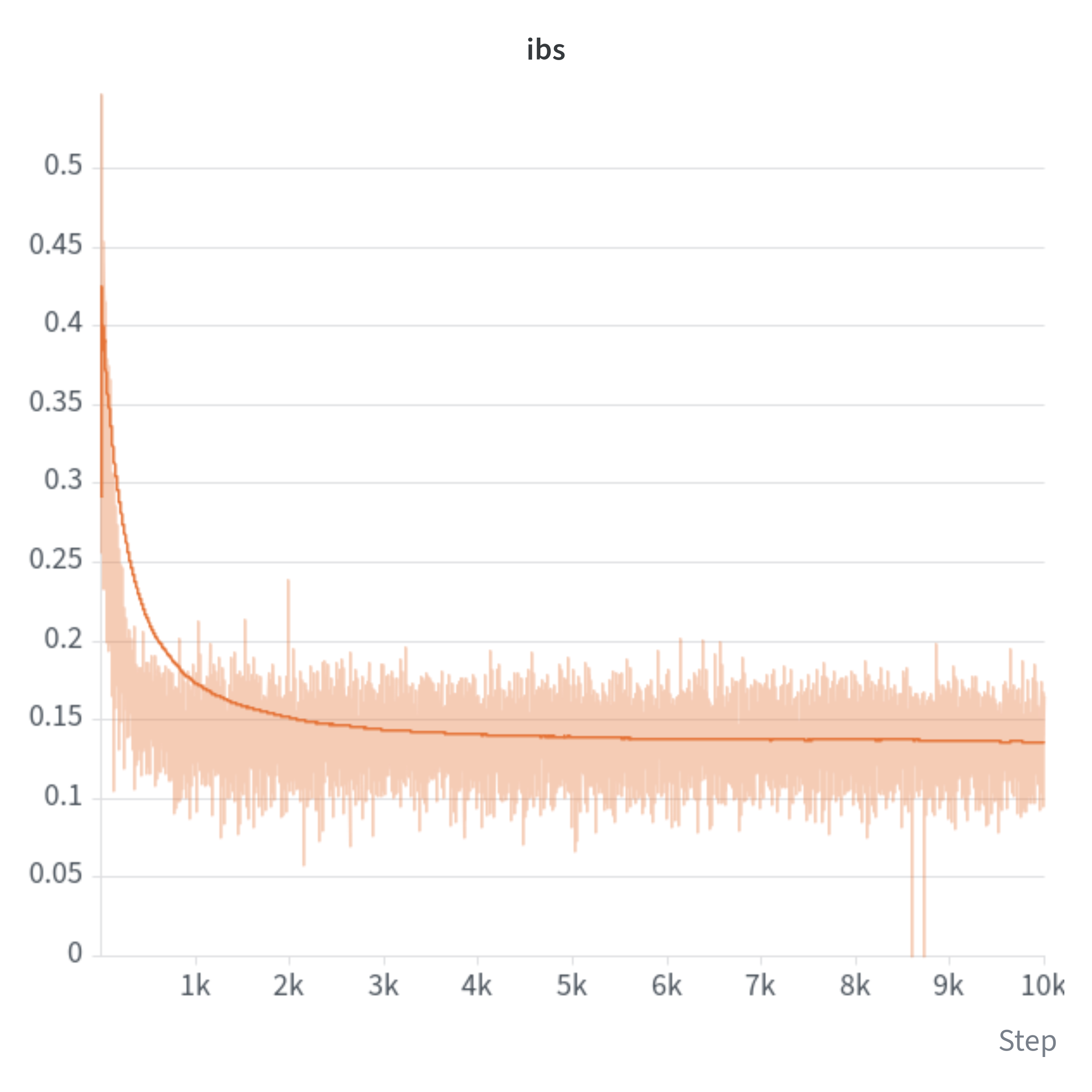}
        \caption{Integrated Brier Score.}
        \label{fig:image2}
    \end{subfigure}
    \hfill
    \begin{subfigure}[b]{0.32\textwidth}
        \centering
        \includegraphics[width=\textwidth]{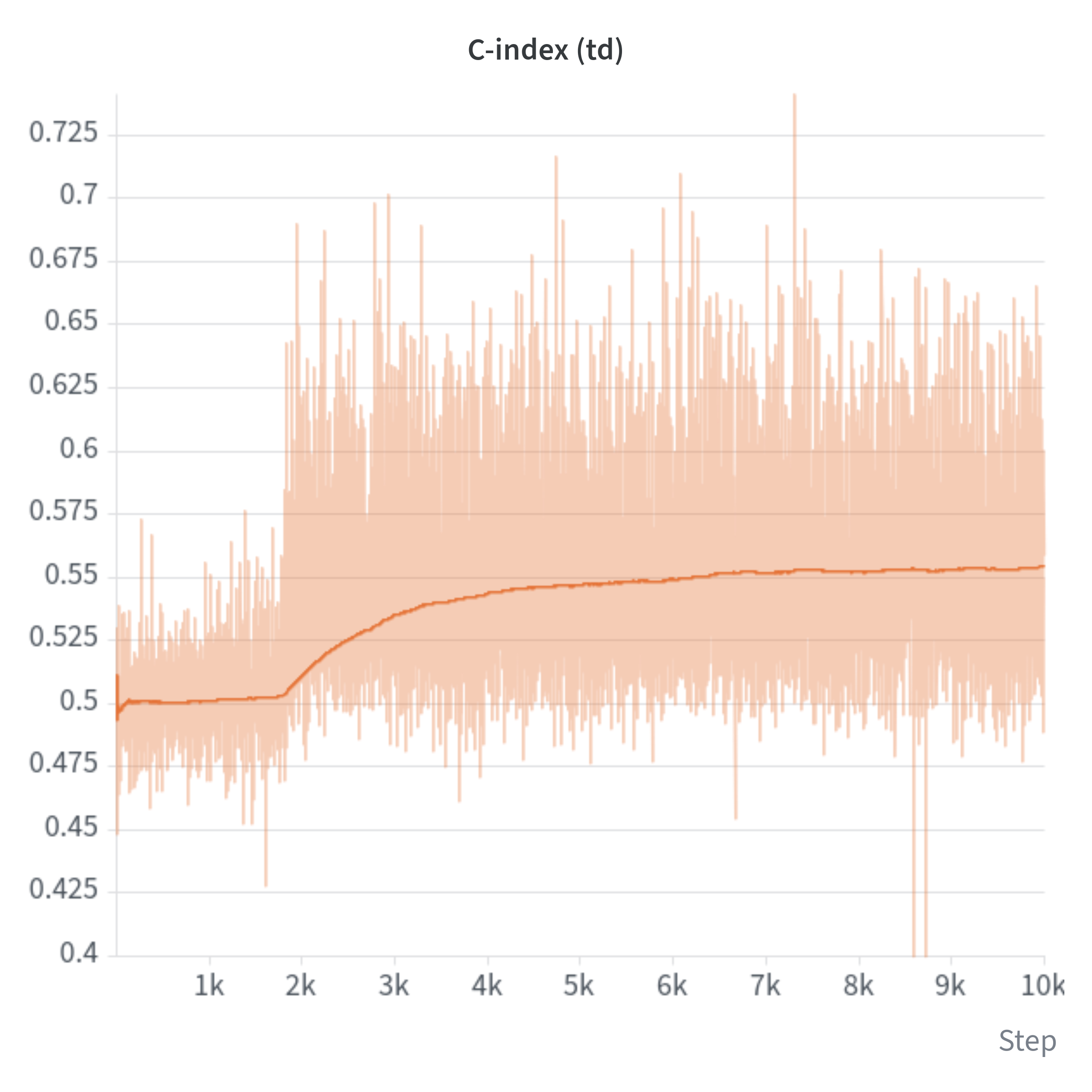}
        \caption{Time-dependent C-index.}
        \label{fig:image3}
    \end{subfigure}
    \caption{Learning curves for the propoded Survival In-Context (SIC) training.}
    \label{fig:trainingcurves}
\end{figure}

As reported in the original TabPFN paper (Appendix F.3 in ~\citet{tabpfnworkshop}), the learning curves are noisy and tend to flatten only after a specific number of datasets, but still remain generally noisy due to their diversity~\footnote{PFNs are trained on millions of synthetic datasets, drawn from a prior and then used as a fixed model at inference time. In each training step, the model consumes a freshly generated dataset that has never been seen before, with substantial variation in structure and difficulty.}. In our setting, we observe the same qualitative behavior, as shown in Figure~\ref{fig:trainingcurves}, with the flattening occurring after 5,000 steps (around 2.5 million datasets) in stage 1, compared to 10 million reported in the original TabPFN. We attribute this to our initialization using classification TabICL, a different training strategy (we follow TabICL), and a larger batch size compared to the original TabPFN strategy. This is also why PFN‑style methods are typically trained for a fixed number of steps in the PFN. 

\section{Datasets}
\label{app:datasets}
We use the datasets from SurvSet~\citep{survset} with the selection criterion: $\#features \ge 3$ and $\#samples \ge 100$. To preprocess raw features, we use TableVectorizer from \textit{skrub}~\citep{skrub}. The statistics about all datasets are provided in Table~\ref{tab:dataset-stats}. For more information about the datasets, we refer to the original SurvSet paper~\citep{survset}. Additionally, we found \texttt{METABRIC}, \texttt{UNOS}, and \texttt{SEER} are often used for benchmarking in Survival Analysis but are not present in SurvSet. The details: 
\begin{itemize} 
    
    \item \texttt{METABRIC} Molecular Taxonomy of Breast Cancer International Consortium. Large breast cancer cohort with detailed clinical data and high‑dimensional gene expression molecular profiles, used for integrative prognostic modeling. Target: Time from diagnosis to breast cancer death. 
    
    \item \texttt{UNOS} United Network for Organ Sharing derived cohort. Large transplant registry data, including recipient, donor, and clinical variables. Target: Time to death after heart transplantation. This dataset has restricted access~\footnote{https://www.unos.org/data/}. 
    
    \item \texttt{SEER} Surveillance, Epidemiology, and End Results Program. Population‑based U.S. cancer registry covering multiple cancer sites with demographics, tumor characteristics, treatments, and vital status. Target: Time from breast cancer diagnosis to death. Competing risk: death from CVD (not included). This dataset has restricted access~\footnote{https://seer.cancer.gov/causespecific/}. 
\end{itemize}

\begin{table}[H]
  \caption{Dataset statistics for 58 real-world datasets used in evaluation (continue in Table~\ref{tab:dataset-stats2}). Raw features are the original covariates before preprocessing by TableVectorizer~\citep{skrub}. 
  Datasets are sorted by number of samples.}
  \label{tab:dataset-stats}
  \centering
  \small
  \begin{tabular}{lrrrr}
    \toprule
    Dataset & \# samples & \# feat. (raw) & \# feat. (proc.) & Censoring \\
    \midrule
    breast & 100 & 4 & 4 & 0.740 \\
    micro.censure & 117 & 40 & 118 & 0.778 \\
    pharmacoSmoking & 125 & 11 & 19 & 0.288 \\
    cgd & 128 & 10 & 25 & 0.656 \\
    veteran & 137 & 6 & 9 & 0.066 \\
    nki70 & 144 & 75 & 77 & 0.667 \\
    stagec & 146 & 7 & 22 & 0.630 \\
    burn & 154 & 11 & 14 & 0.688 \\
    d.oropha.rec & 192 & 12 & 30 & 0.276 \\
    wpbc & 198 & 32 & 32 & 0.763 \\
    Melanoma & 205 & 5 & 5 & 0.722 \\
    hepatoCellular & 227 & 43 & 43 & 0.573 \\
    cancer & 228 & 8 & 30 & 0.276 \\
    mgus & 241 & 9 & 13 & 0.066 \\
    e1684 & 284 & 3 & 3 & 0.310 \\
    pbc & 312 & 6 & 6 & 0.599 \\
    Pbc3 & 349 & 12 & 21 & 0.825 \\
    ova & 358 & 5 & 13 & 0.257 \\
    diabetes & 394 & 4 & 4 & 0.607 \\
    retinopathy & 394 & 7 & 12 & 0.607 \\
    zinc & 431 & 13 & 23 & 0.812 \\
    phpl04K8a & 442 & 21 & 21 & 0.466 \\
    Unemployment & 452 & 5 & 8 & 0.434 \\
    whas500 & 461 & 16 & 18 & 0.618 \\
    prostate & 502 & 15 & 28 & 0.295 \\
    cost & 518 & 13 & 13 & 0.220 \\
    follic & 541 & 5 & 4 & 0.216 \\
    vlbw & 617 & 22 & 55 & 0.827 \\
    uis & 628 & 8 & 17 & 0.191 \\
    GBSG2 & 686 & 8 & 10 & 0.564 \\
    dataOvarian1 & 912 & 160 & 163 & 0.404 \\
    colon & 929 & 9 & 14 & 0.513 \\
    grace & 1{,}000 & 5 & 5 & 0.676 \\
    rdata & 1{,}040 & 4 & 7 & 0.474 \\
    LeukSurv & 1{,}043 & 7 & 30 & 0.157 \\
    actg & 1{,}151 & 11 & 20 & 0.917 \\
    dataDIVAT2 & 1{,}837 & 4 & 4 & 0.683 \\
    TRACE & 1{,}878 & 6 & 6 & 0.490 \\
    METABRIC & 1{,}904 & 9 & 9 & 0.421 \\
    scania & 1{,}931 & 5 & 9 & 0.438 \\
    Aids2 & 2{,}839 & 4 & 14 & 0.380 \\
    rott2 & 2{,}982 & 11 & 13 & 0.573 \\
    UnempDur & 3{,}241 & 6 & 6 & 0.387 \\
    divorce & 3{,}371 & 3 & 5 & 0.694 \\
    \bottomrule
  \end{tabular}
\end{table}

\begin{table}[H]
  \caption{Table~\ref{tab:dataset-stats} continued. Dataset statistics for 58 real-world datasets used in evaluation. Raw features are the original covariates before preprocessing by TableVectorizer~\citep{skrub}. 
  Datasets are sorted by number of samples.}
  \label{tab:dataset-stats2}
  \centering
  \small
  \begin{tabular}{lrrrr}
    \toprule
    Dataset & \# samples & \# feat. (raw) & \# feat. (proc.) & Censoring \\
    \midrule
    acath & 3{,}504 & 3 & 3 & 0.334 \\
    smarto & 3{,}873 & 26 & 39 & 0.881 \\
    nwtco & 4{,}028 & 7 & 10 & 0.858 \\
    dataDIVAT3 & 4{,}267 & 7 & 17 & 0.944 \\
    Framingham & 4{,}699 & 7 & 18 & 0.687 \\
    rhc & 5{,}735 & 53 & 77 & 0.351 \\
    dataDIVAT1 & 5{,}943 & 5 & 17 & 0.836 \\
    Dialysis & 6{,}805 & 4 & 37 & 0.764 \\
    flchain & 7{,}874 & 10 & 34 & 0.725 \\
    support2 & 9{,}105 & 35 & 73 & 0.319 \\
    prostateSurvival & 14{,}294 & 3 & 8 & 0.944 \\
    hdfail & 52{,}422 & 6 & 38 & 0.945 \\
    UNOS & 62{,}645 & 47 & 104 & 0.502 \\
    SEER & 280{,}845 & 24 & 24 & 0.876 \\
    \bottomrule
  \end{tabular}
\end{table}

\section{Evaluation Metrics}
\label{app:metrics}
\paragraph{Time-dependent concordance index.} td-CI~\citep{antolini05} is a common evaluation that quantifies the ability of a model to order the relative risks pairwise:  

\begin{equation}
\text{td-CI} = \mathbb{P}(\widehat{S}(t_i|x_i) < \widehat{S}(t_i|x_j)| t_i< t_j, 
\end{equation}

We use the pycox ~\citep{pycox} \textit{adjusted} Antolini estimator (default option).

\paragraph{Integrated Brier Score.} $IBS$ is the integration of the Brier score
across all time points on the interval $[t_{min}, t_{max}]$:
\begin{equation}
    IBS = \frac{1}{t_{max}-t_{min}} \int_{t_{min}}^{t_{max}}BS(u)du
\end{equation}
The score is non-negative. The lower the better.
The Brier Score (BS) evaluates the mean squared error between the predicted survival curve with step function of the observed event. Namely, for $\widehat{S}_{\text {censor }}$ obtained with KM-estimator:
\begin{equation}
\begin{aligned}
&\mathrm{BS}(t):=\frac{1}{n} \sum_{i=1}^n\left[\frac{\widehat{S}\left(t \mid x_i\right)^2 e_i \mathbf{1}\left\{t_i \leq t\right\}}{\widehat{S}_{\text {censor }}\left(t_i\right)}+\frac{\left(1-\widehat{S}\left(t \mid x_i\right)\right)^2 \mathbf{1}\left\{t_i>t\right\}}{\widehat{S}_{\text {censor }}(t)}\right]\\
\end{aligned}
\end{equation}
To avoid numerically unstable values in the far right tail (see ~\citet{chen26}), we report IBS on a dense 100-bin grid up to the 90th percentile of train follow-up times. 

\paragraph{Distributional Calibration.} D-Cal~\citep{haider20} assesses how well predicted survival probabilities align with observed outcomes based on a goodness-of-fit test. D-Cal discretizes the predicted survival probabilities at the true event times into $n$ equidistant intervals in $[0,1]$, and performs a chi-squared test for the uniformity of the distribution. As suggested in the original paper~\citep{haider20}, we set $n=10$ and report the
number of folds with $p<0.05$, representing the number of misscalibrated folds in the dataset.

\section{Details on Hyperparameter Tuning for Baseline Comparison}
\label{app:hyperparams}
To offer a fair comparison, we extensively tune all baselines. Our choice of hyperparameter grid is based on the~\citep{erickson25, chen24}. We conduct the hyperparameter search space over the grid in Table~\ref{tab:hyparams} with the termination condition: 50 trials or 5h per fold (all started trials are completed if they exceed the 5h constraint, therefore, the total training time can exceed 25h for 5 folds). The default values are in brackets and included as the start configuration in the trial 1. For the larger datasets with over 10,000 samples (\texttt{prostateSurvival, hdfail, UNOS, SEER}), we extend the depth parameters to [1,6] for the neural net approaches. 

For the baseline models, we use the following software implementations: \textit{lifelines} library~\citep{lifelines} for CoxPH; \textit{sksurv} library~\citep{sksurv} for RSF, \textit{xgboost} library~\citep{xgboost} for XGB, \textit{pycox} library~\citep{pycox} for DeepSurv and DeepHit, the Jeanselme's implementation~\footnote{\url{https://github.com/Jeanselme/SuMo-net}} of SuMo-Net~\citep{lee24}, the original implementation of TabICL~\citep{qu25} for TabICL + Stacking, the original implementation of MITRA\_CF~\citep{kim26r}, and the original implementation of SurvivalPFN~\citep{qi26}.

All time-measurement experiments are executed on the same type of resource, with 32 CPU cores, 256GB RAM, and one A40 49 GB GPU (CUDA accelerator used for DeepSurv, DeepHit, SuMo-Net, TabICL + Stacking, MITRA\_CF, SurvPFN, SurvivalPFN). We seek to facilitate fair comparison in real-world settings. Therefore, we use parallelization in Optuna to speed up the baseline training. We use n\_jobs=16 for all baselines as default. If the baseline fails (due to exceeding 256GB RAM), we switch to lower number of n\_jobs, namely for RSF n\_jobs=8 for \texttt{dataDIVAT1, Dialysis, flchain, rhc, support2} and n\_jobs=1 for \texttt{prostateSurvival, hdfail, UNOS, SEER}.

\begin{table}[H]
\centering
\caption{Summary of hyperparameter search spaces}
\label{tab:hyparams}
\begin{tabular}{lll}
\hline
Model & Hyperparameter (default) & Domain \\
\toprule
Cox~\citep{cox72} & l1\_ratio (0.0) & [0.0, 1.0] \\
 & penalizer (0.0) & [1e{-}6, 1] \\
 \midrule
DeepSurv~\citep{katzman18} & optimizer (adam) & \{adam\} \\
 & weight\_decay (0.001) & [1e{-}6, 1e{-}2] \\
 & momentum (0.0) & [0.0, 0.9] \\
 & lr (0.001) & [1e{-}5, 1e{-}2] \\
 & dropout (0.0) & [0.0, 0.5] \\
 & layers ([32, 32]) & layers = [
    [width] * depth \\ & & 
    for width in \{8,16,32,64,128,256\} \\ & & 
    for depth in [1, 3]
] \\
\midrule
DeepHit~\citep{lee18} & alpha (0.2) & [0.1, 0.9] \\
 & sigma (0.1) & [0.1, 0.9] \\
 & optimizer (adam) & \{adam\} \\
 & weight\_decay (0.001) & [1e{-}6, 1e{-}2] \\
 & momentum (0.0) & [0.0, 0.9] \\
 & lr (0.001) & [1e{-}5, 1e{-}2] \\
 & dropout (0.0) & [0.0, 0.5] \\
 & num\_durations (10) & [10, 100] \\
 & batch\_size (256) & \{32, 64, 128, 256, 512\} \\
 & layers ([32, 32]) &  layers = [
    [width] * depth \\ & & 
    for width in \{8,16,32,64,128,256\} \\ & & 
    for depth in [1, 3]
    ]
 \\
\midrule
XGBoost~\citep{chen16} & max\_feat\_sqrt\_fac (1.0)& [0.5, 2.0] \\
 & eta (0.3) & [0.01, 1.0] \\
 & num\_parallel\_tree (1) & [1, 20] \\
 & subsample (1.0) & [0.1, 1.0] \\
 & max\_depth (6) & [1, 20] \\
\midrule
RSF~\citep{ishawaran08} &n\_estimators (100) & [10, 1000] \\
 & max\_depth (None) & [1, 20] \\
 & max\_feat\_sqrt\_fac (1.0) & [0.5, 2.0] \\

\midrule
SuMo-Net~\citep{Rindt21} & optimizer (adam) & \{adam\} \\
 & weight\_decay (0.001) & [1e{-}6, 1e{-}2] \\
 & momentum (0.0) & [0.0, 0.9] \\
 & lr (0.001) & [1e{-}5, 1e{-}2] \\
 & dropout (0.0) & [0.0, 0.5] \\
 & batch\_size (256) & \{32, 64, 128, 256, 512\} \\
 & layers ([32, 32]) &  layers = [
    [width] * depth \\ & & 
    for width in \{8,16,32,64,128,256\} \\ & & 
    for depth in [1, 3]
    ] \\
 & layers\_surv ([32]) &  layers\_surv = [
    [width] * depth \\ & & 
    for width in \{8,16,32,64,128,256\} \\ & & 
    for depth in [1, 3]
    ]
 \\
 \bottomrule
\end{tabular}
\end{table}

\section{Convergence Failures and Restrictions of Baselines}
\label{app:failures}

\subsection{CoxPH (default) convergence failures.} 
\label{app:coxph-default} 
CoxPH (default) is the unpenalized lifelines~\citep{lifelines} Cox model ($penalizer=0$, $l1\_ratio=0$). This failed to converge on 41 of 58 datasets, typically with lifelines reporting matrix inversion problems due to high collinearity or a singular matrix. Because fitting fails, no survival predictions are produced, and all metrics (C-index, IBS, and D-cal) are undefined on those datasets. In Elo, such outcomes are counted as losses against opponents with finite scores. In practice, either covariate preprocessing, such as PCA on raw features~\citep{apollo}, or regularization can mitigate these failures. We include a tuned CoxPH variant that searches over elastic-net regularization ($penalizer$ and $l1\_ratio$) and intentionally keep CoxPH (default) unregularized.

The list of the failed datasets: \texttt{Aids2}, \texttt{Dialysis}, \texttt{Framingham}, \texttt{GBSG2}, \texttt{LeukSurv}, \texttt{Pbc3}, \texttt{SEER}, \texttt{Unemployment}, \texttt{actg}, \texttt{burn}, \texttt{cancer}, \texttt{cgd}, \texttt{colon}, \texttt{d.oropha.rec}, \texttt{dataDIVAT1}, \texttt{dataDIVAT3}, \texttt{dataOvarian1}, \texttt{divorce}, \texttt{flchain}, \texttt{hdfail}, \texttt{mgus}, \texttt{micro.censure}, \texttt{nki70}, \texttt{nwtco}, \texttt{ova}, \texttt{pharmacoSmoking}, \texttt{prostate}, \texttt{prostateSurvival}, \texttt{rdata}, \texttt{retinopathy}, \texttt{rhc}, \texttt{rott2}, \texttt{scania}, \texttt{smarto}, \texttt{stagec}, \texttt{support2}, \texttt{uis}, \texttt{veteran}, \texttt{vlbw}, \texttt{whas500}, \texttt{zinc}.

\subsection{Other baseline failures and restrictions.}
\label{app:otherfailures}
The following baselines failed to produce predictions:
\begin{itemize}
    \item RSF on \texttt{UNOS}. Out of Memory (exceeds 256GB RAM) due to a high number of unique times and n\_estimators.
    \item TabICL+Stacking on \texttt{UNOS}. Out of Memory (exceeds 256GB RAM) due to the high number of unique times.
\end{itemize}
The following baselines require a restricted context size:
\begin{itemize}
    \item SurvPFN was originally evaluated with a context of up to 1,000 samples. On our 49\,GB GPU, it runs out of memory beyond 2,000 samples, so we cap the context at 2,000 samples, selected by random subsampling.
    \item TabIC+Stacking is capped at the recommended maximum of 100,000 context samples, since larger contexts exceed the 49\,GB of GPU memory. The context is selected by random subsampling.
\end{itemize}

\section{Additional Results}

\subsection{Distributional Calibration}
\label{app:calibration} 

\begin{figure}[H]

    \centerline{\includegraphics[width=\textwidth]{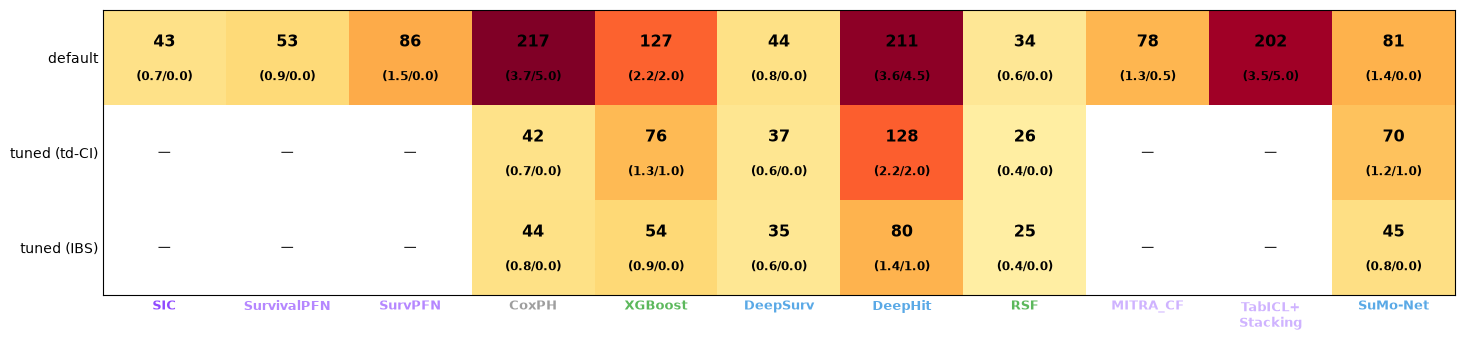}}
    \caption{Distributional Calibration (D-cal) performance (total (mean/median) number of misscalibrated folds per dataset) across 58 (with 5 folds each) real-world datasets. The Survival In-context (SIC) is a proposed method compared to the classical linear (CoxPH\citep{cox72}), neural net (SuMo-Net\citep{Rindt21}, DeepHit\citep{lee18}, DeepSurv\citep{katzman18}), tree-based (RSF\citep{ishawaran08}, XGB\citep{chen16}) baselines; naive Prior-fitted Network (PFN) TabICL\citep{qu25} + Stacking\citep{craig21} baseline and the concurrent PFN baselines MITRA\_CF~\citep{kim26r} (classification prior), SurvPFN~\citep{boehm26} (survival prior), and SurvivalPFN~\citep{qi26} (survival prior).}
        \label{fig:dcal}

\end{figure}

\subsection{Additional Metrics}
\label{app:allresults}

\begin{figure}[H]
        \centering
        \includegraphics[width=\linewidth]{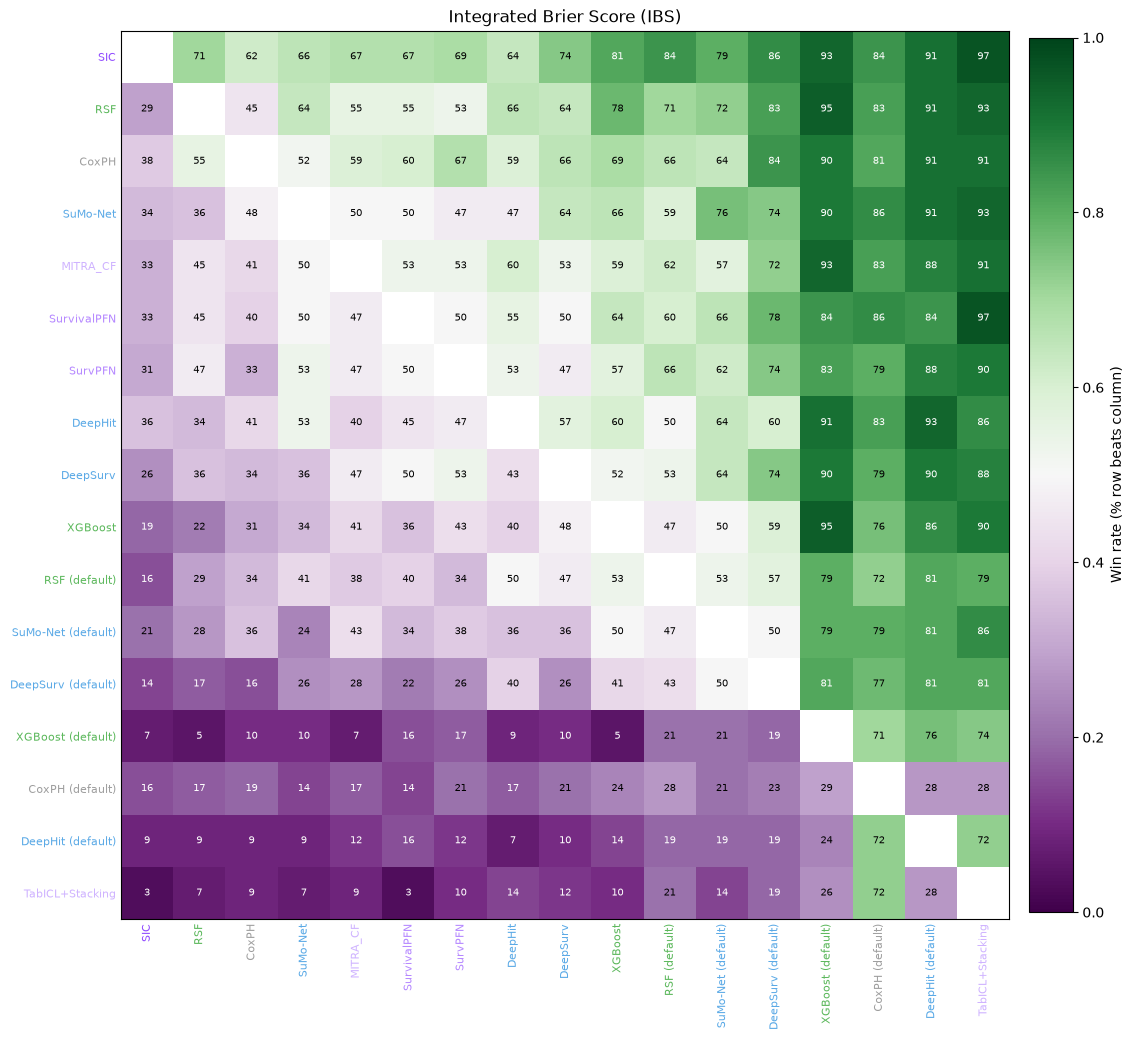}
           \caption{
    Pairwise-winrate matrix for Integrated Brier Score (IBS): percentage of row beats column method. SIC is the proposed method. 
    }
        \label{fig:ibswinrate}
    \end{figure}

\begin{figure}[H]
        \centering
        \includegraphics[width=\linewidth]{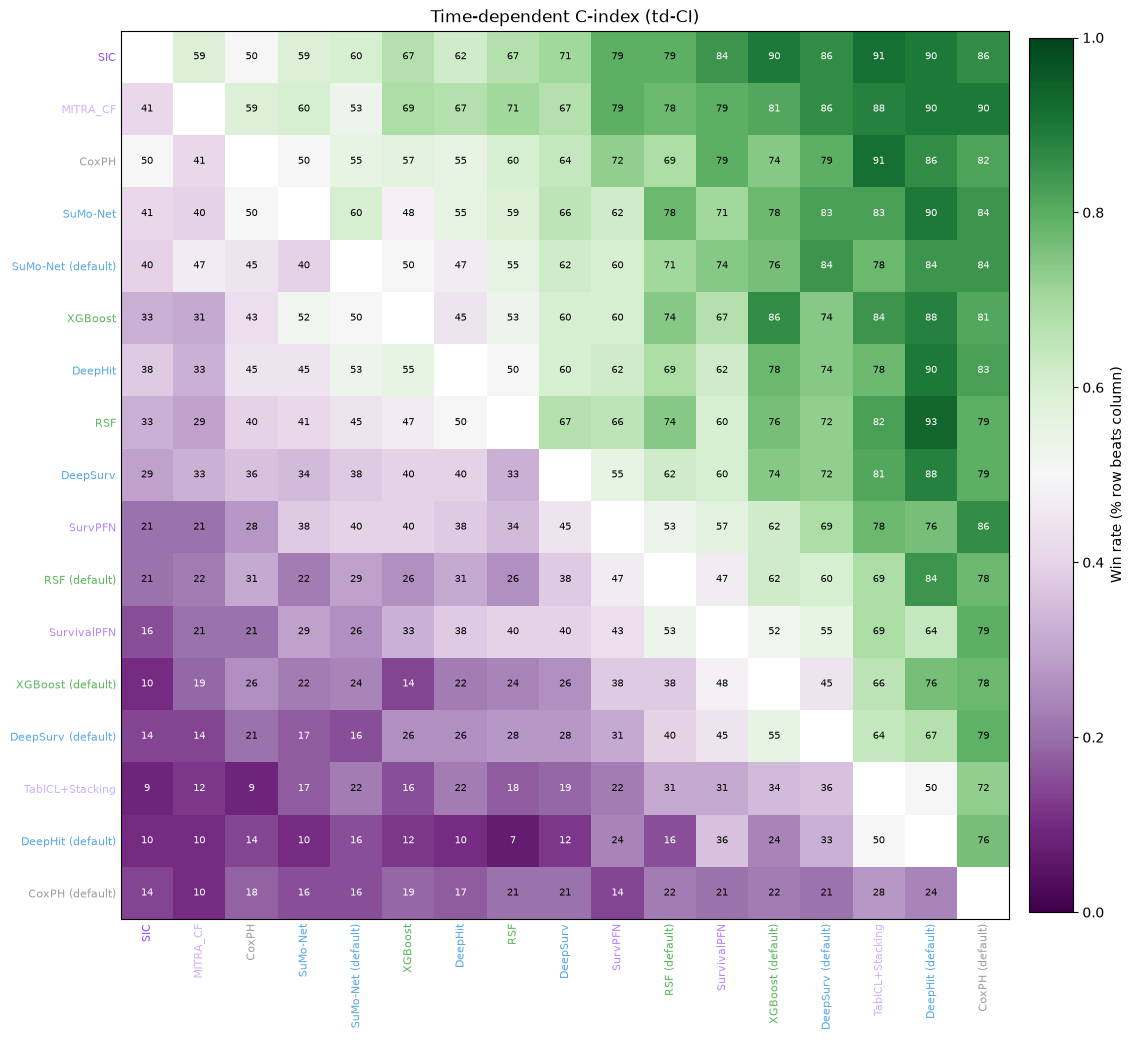}
            \caption{
    Pairwise-winrate matrix for time-dependent C-index (td-CI): percentage of row beats column method. SIC is the proposed method. 
    }
        \label{fig:tdciwinrate}
    \end{figure}

\begin{figure}[H]
    \centering

    \begin{subfigure}{\columnwidth}
        \centering
        \includegraphics[width=\linewidth]{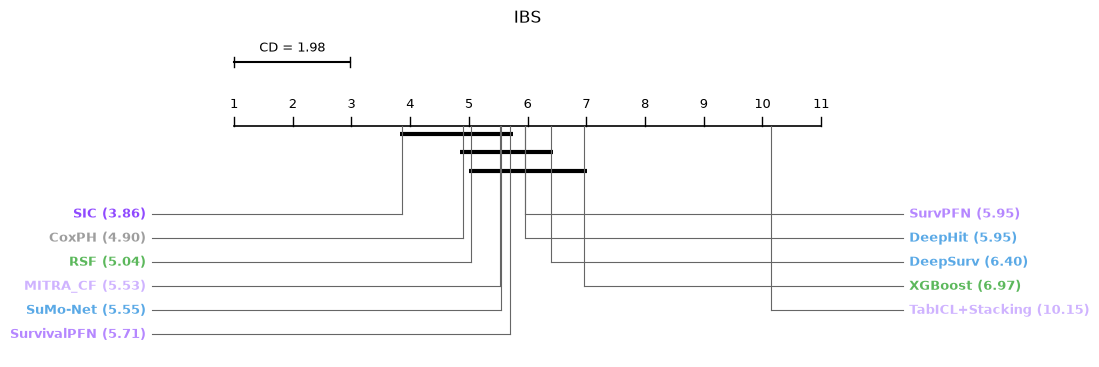}
        \label{fig:ibscd}
    \end{subfigure}

    \vspace{0.5em}

    \begin{subfigure}{\columnwidth}
        \centering
        \includegraphics[width=\linewidth]{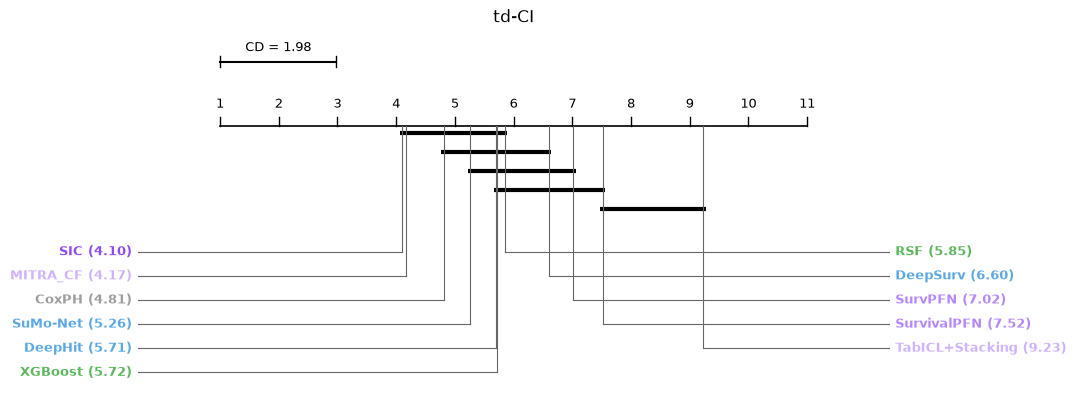}
        \label{fig:tdcicd}
    \end{subfigure}

    \caption{
    Critical difference diagram for (upper) Integrated Brier Score (IBS) and (lower) time-dependent C-index (td-CI). SIC is a proposed method.
    }
    \label{fig:criticaldiff}
\end{figure}

For every method and dataset we report the mean over the outer cross-validation folds of the IBS and the td-CI. All aggregate statistics below are computed from these per-dataset means, so every dataset contributes equally. When a method fails to produce a finite score on a dataset, that outcome is counted as a loss against any opponent with a finite score (see further details in Appendix~\ref{app:otherfailures}). We follow ~\citet{erickson25} to calculate the following metrics in Tables~\ref{tab:elo_extras_ibs},~\ref{tab:elo_extras_td_ci} and Figures~\ref{fig:ibswinrate},~\ref{fig:tdciwinrate},~\ref{fig:criticaldiff}:
\begin{itemize}
    \item \textbf{Elo} On each dataset, every pair of methods plays one battle, won by the method with the better per-dataset mean; exact ties count as half a win for each side. Ratings use the chess convention (base 10, scale 400), so a gap of 400 Elo corresponds to a 10:1 expected win rate; Elo depends only on the direction of differences, not on their magnitude. We anchor the scale by fixing CoxPH (default) at 1000. 95\% confidence intervals are obtained from 200 bootstrap rounds that resample datasets with replacement, refit and re-anchor the ratings in every round, and take the 2.5\% and 97.5\% percentiles; the anchor therefore has no interval.

    \item \textbf{Pairwise win rate.} For an ordered pair $(A,B)$, the fraction of datasets on which $A$ beats $B$, with ties counted as one half and forfeits as above.

    \item \textbf{Win rate.} The fraction of all battles a method wins across all opponents and datasets (ties counted as one half), i.e.\ the mean of its row of the pairwise win-rate matrix.

    \item \textbf{Improvability.} With $\mathrm{err}$ denoting IBS or $1-\text{td-CI}$, the improvability of a method on dataset $i$ is $(\mathrm{err}_i - \mathrm{err}^{\star}_i)/\mathrm{err}_i \times 100\%$, where $\mathrm{err}^{\star}_i$ is the best error of any compared method on that dataset; we report the mean over datasets. It ranges from 0\% (best on every dataset) to 100\% and, unlike Elo, is sensitive to the size of the gaps. A failed result counts as 100\%.

    \item \textbf{Mean and median rank.} On each dataset, methods are ranked by their per-dataset mean (1 = best).

    \item \textbf{Critical difference diagram} is computed with a Friedman test on the per-dataset ranks of the tuned methods (without (default)) with significance level $\alpha = 0.05$.

\end{itemize}

\begin{table}[H]
\centering
\caption{Integrated Brier Score (IBS): Elo ratings (95\% CI; CoxPH (default) pinned at 1000), mean / median per-dataset rank over all entries (failed results rank last), pairwise win rate and improvability, sorted by Elo. SIC is a proposed method. Best per column in \textbf{bold}, second best in \textit{italics}.}
\label{tab:elo_extras_ibs}
\small
\begin{tabular}{lccccc}
\toprule
Method & Elo [95\% CI] $\uparrow$ & Mean rank $\downarrow$ & Median rank $\downarrow$ & Win rate (\%) $\uparrow$ & Improv. (\%) $\downarrow$ \\
\midrule
SIC & \textbf{1480} {\scriptsize [1385, 1617]} & \textbf{4.67} & \textbf{4.0} & \textbf{77.3} & \textbf{3.4} \\
RSF & \textit{1402} {\scriptsize [1293, 1554]} & 6.15 & \textit{6.0} & \textit{68.5} & 5.6 \\
CoxPH & 1399 {\scriptsize [1305, 1529]} & \textit{6.14} & \textit{6.0} & 68.2 & 4.8 \\
SuMo-Net & 1358 {\scriptsize [1255, 1482]} & 6.76 & 6.5 & 63.1 & 5.4 \\
MITRA\_CF & 1350 {\scriptsize [1259, 1483]} & 6.97 & \textit{6.0} & 62.2 & \textit{4.4} \\
SurvivalPFN & 1347 {\scriptsize [1262, 1458]} & 7.12 & 7.0 & 61.7 & 6.4 \\
SurvPFN & 1332 {\scriptsize [1230, 1474]} & 7.43 & 6.5 & 59.9 & 5.8 \\
DeepHit & 1324 {\scriptsize [1205, 1468]} & 7.47 & 8.0 & 58.8 & 6.1 \\
DeepSurv & 1311 {\scriptsize [1211, 1424]} & 7.93 & 8.0 & 57.2 & 5.2 \\
XGBoost & 1264 {\scriptsize [1148, 1389]} & 8.84 & 10.0 & 51.1 & 7.2 \\
RSF (default) & 1258 {\scriptsize [1124, 1406]} & 8.93 & 8.5 & 50.3 & 6.1 \\
SuMo-Net (default) & 1240 {\scriptsize [1133, 1374]} & 9.26 & 9.5 & 48.1 & 9.0 \\
DeepSurv (default) & 1191 {\scriptsize [1097, 1311]} & 10.39 & 11.0 & 41.7 & 10.6 \\
XGBoost (default) & 1028 {\scriptsize [908, 1149]} & 13.24 & 14.0 & 23.6 & 13.9 \\
CoxPH (default) & 1000 {\scriptsize [1000, 1000]} & 13.66 & 17.0 & 20.9 & 71.5 \\
DeepHit (default) & 997 {\scriptsize [870, 1134]} & 13.69 & 15.0 & 20.7 & 22.7 \\
\makecell[l]{TabICL+\\Stacking} & 948 {\scriptsize [819, 1075]} & 14.35 & 16.0 & 16.5 & 38.1 \\
\bottomrule
\end{tabular}
\end{table}

\begin{table}[H]
\centering
\caption{Time-dependent C-inted (td-CI): Elo ratings (95\% CI; CoxPH (default) pinned at 1000), mean / median per-dataset rank over all entries (failed results rank last), pairwise win rate and improvability, sorted by Elo. SIC is a proposed method. Best per column in \textbf{bold}, second best in \textit{italics}.}
\label{tab:elo_extras_td_ci}
\small
\begin{tabular}{lccccc}
\toprule
Method & Elo [95\% CI] $\uparrow$ & Mean rank $\downarrow$ & Median rank $\downarrow$ & Win rate (\%) $\uparrow$ & Improv. (\%) $\downarrow$ \\
\midrule
SIC & \textbf{1453} {\scriptsize [1364, 1592]} & \textbf{5.21} & \textit{4.5} & \textbf{73.8} & \textit{5.3} \\
MITRA\_CF & \textit{1441} {\scriptsize [1348, 1565]} & \textit{5.40} & \textbf{4.0} & \textit{72.4} & \textbf{5.1} \\
CoxPH & 1392 {\scriptsize [1296, 1547]} & 6.37 & 5.5 & 66.6 & 6.2 \\
SuMo-Net & 1383 {\scriptsize [1289, 1530]} & 6.52 & 6.0 & 65.4 & 7.2 \\
SuMo-Net (default) & 1358 {\scriptsize [1257, 1500]} & 6.97 & 6.5 & 62.3 & 9.6 \\
XGBoost & 1351 {\scriptsize [1249, 1484]} & 7.19 & 6.0 & 61.4 & 7.7 \\
DeepHit & 1347 {\scriptsize [1227, 1494]} & 7.19 & 7.0 & 60.9 & 8.3 \\
RSF & 1338 {\scriptsize [1237, 1493]} & 7.46 & 7.0 & 59.7 & 9.2 \\
DeepSurv & 1291 {\scriptsize [1180, 1434]} & 8.47 & 8.0 & 53.4 & 8.7 \\
SurvPFN & 1258 {\scriptsize [1168, 1369]} & 9.16 & 9.5 & 49.0 & 11.8 \\
RSF (default) & 1214 {\scriptsize [1119, 1364]} & 10.07 & 11.0 & 43.3 & 10.3 \\
SurvivalPFN & 1207 {\scriptsize [1115, 1324]} & 10.24 & 10.5 & 42.3 & 15.0 \\
XGBoost (default) & 1157 {\scriptsize [1056, 1274]} & 11.22 & 12.0 & 36.0 & 13.9 \\
DeepSurv (default) & 1153 {\scriptsize [1070, 1270]} & 11.39 & 11.0 & 35.5 & 19.5 \\
\makecell[l]{TabICL+\\Stacking} & 1074 {\scriptsize [964, 1210]} & 12.75 & 15.0 & 26.3 & 20.5 \\
DeepHit (default) & 1038 {\scriptsize [924, 1164]} & 13.41 & 15.0 & 22.5 & 19.4 \\
CoxPH (default) & 1000 {\scriptsize [1000, 1000]} & 14.00 & 17.0 & 18.9 & 72.4 \\
\bottomrule
\end{tabular}
\end{table}

\subsection{Performance with respect to the number of samples.}

\label{app:rank}
\begin{figure}[H]
    \centering

    \begin{subfigure}{\columnwidth}
        \centering
        \includegraphics[width=\linewidth]{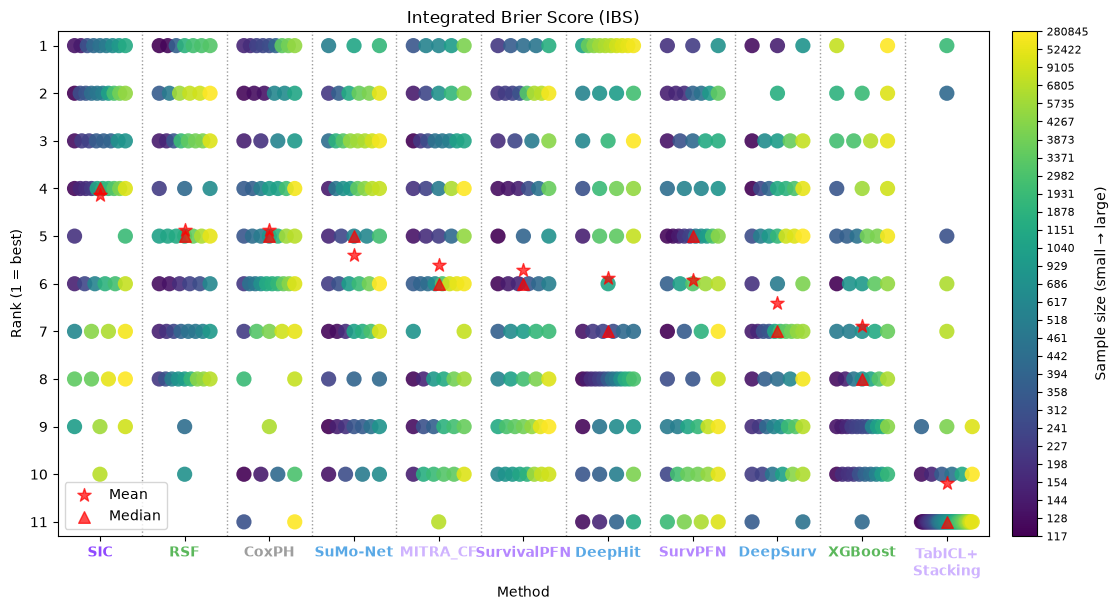}
        \label{fig:ibsrank}
    \end{subfigure}

    \vspace{0.5em}

    \begin{subfigure}{\columnwidth}
        \centering
        \includegraphics[width=\linewidth]{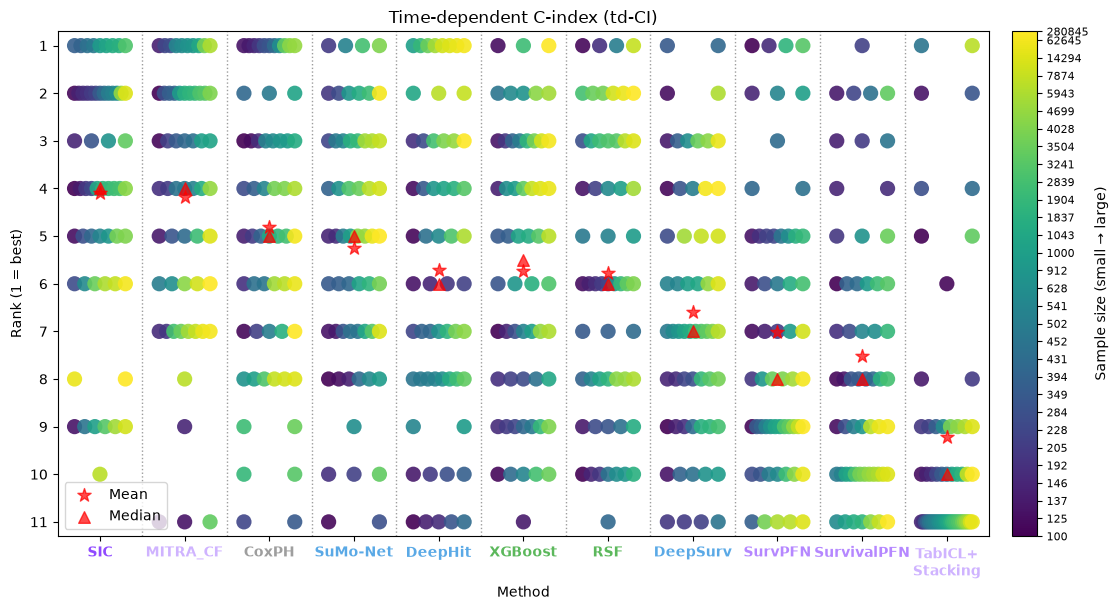}
        \label{fig:tdcirank}
    \end{subfigure}

    \caption{
    Dataset-wise method ranks (best = 1) without default methods across 58 datasets, ordered by dataset size, evaluated using (upper) Integrated Brier Score (IBS) and(lower) time-dependent C-index (td-CI).
    The proposed Survival In-context (SIC) method attends higher ranks on the small- and medium-sized datasets (darker colors) and lower ranks on the larger datasets (lighter colors). 
    }
    \label{fig:rank}
\end{figure}

\section{Details of Simulation Experiment}
\label{app:simulation}
 We simulate six survival datasets: (Exp-PH-100) $(x_1, x_2, x_3) \sim N(0, I^3)$ with Exponential-PH assumptions for survival times, independent 50\% censoring, and 100 samples; (Exp-PH-1000) and (Exp-PH-10000) as (Exp-PH-100) but with 1000 and 10000 samples, respectively; (DAG-Lognormal-AFT-100/1000/10000) Simple Non-linear DAG-dependency $x_1 \sim N(0,1), x2 \sim N(0, 1), x_3 \sim N(2*x_1*x_2, 1), \eta \sim N(x_1 + x_2^2 + \sqrt{|x_3|}, 1)$ for Lognormal-AFT survival times generation, with 100/1000/10000 samples.

 We apply the same evaluation protocol as in the main experiment: nested 5-fold cross-validation, with models hyperparameter-optimized for IBS to compare in IBS and td-CI to compare in td-CI. All time-measurement experiments are executed on the same type of resource, with 32 CPU cores, 256GB RAM, and one A40 49 GB GPU (CUDA accelerator used for SuMo-Net). We use parallelization in Optuna to speed up the baseline training. We use n\_jobs=16 for all baselines. Due to exceeding 256GB RAM, RSF fails on 10,000 samples. We switch to n\_jobs=1 to succeed in the training.

\section{Ablations}
\label{app:ablations}

\subsection{Priors}
\label{app:ablation:prior}
\begin{figure}[H]
  \centering
\includegraphics[width=\textwidth]{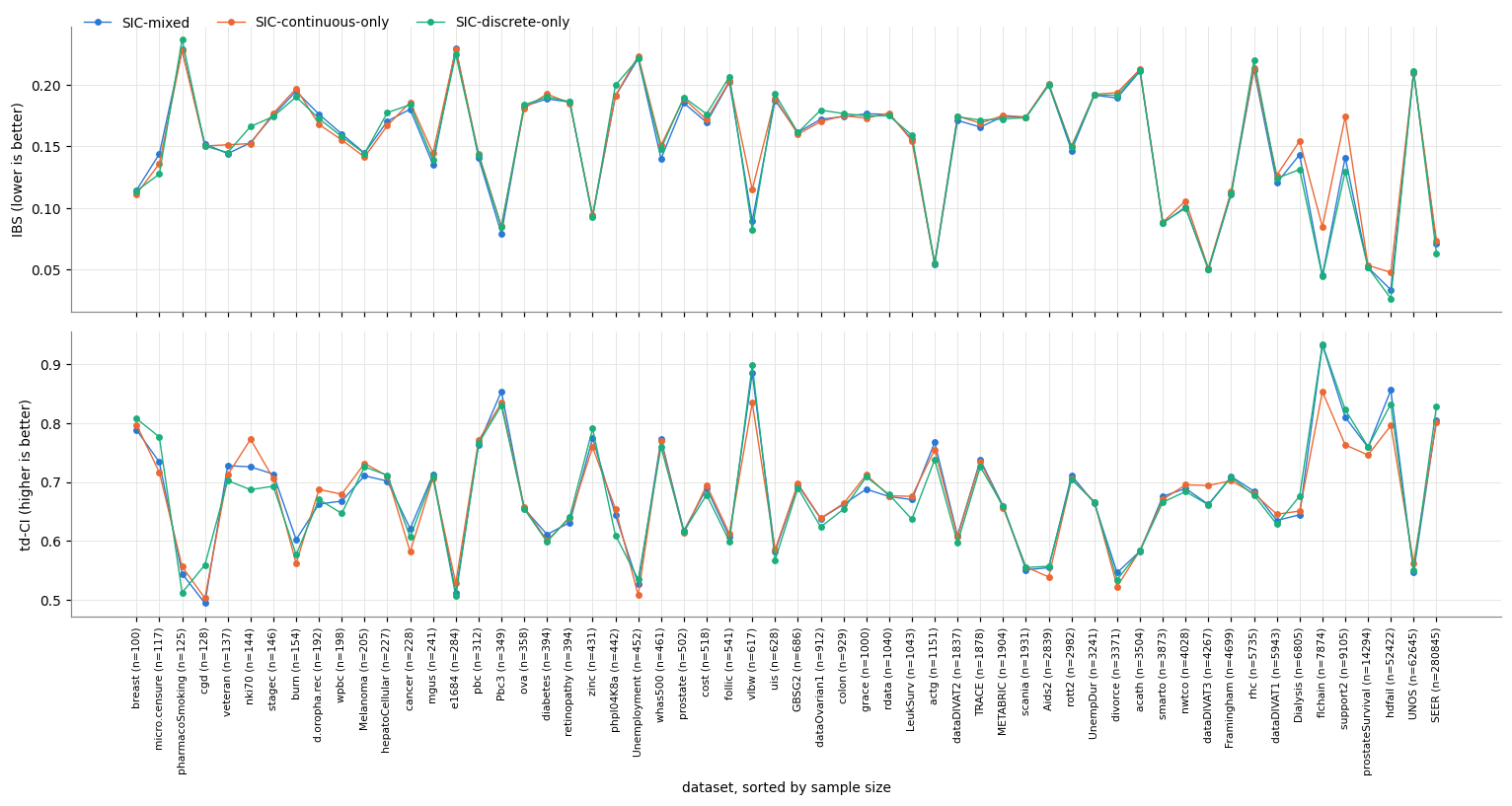}
    \caption{Performance in (upper) Integrated Brier Score (IBS) and (lower) time-dependent C-index (td-CI) across 58 datasets in stage 1 after 5000 steps for different \textit{Survival Priors}: Discrete-only, Continuous-only, and Mixed of both (main option used in evaluation).}
    \label{fig:priors}
\end{figure}

\begin{table}[H]
\centering\small
\caption{Pairwise win-rate comparison on the 58 real-world datasets in stage 1 after 5000 steps for different \textit{Survival Priors}: Discrete-only, Continuous-only, and Mixed of both (main option used in evaluation). Win rate of model A over model B across datasets. }
\label{tab:abl_prior}
\begin{tabular}{llcc}
\toprule
Model A & Model B & td-CI & IBS \\
\midrule
SIC-mixed & SIC-discrete-only & 0.60 & 0.59 \\
SIC-mixed & SIC-continuous-only & 0.48 & 0.72 \\
\bottomrule
\end{tabular}
\end{table}

\subsection{Curriculum Learning}
\label{app:ablation:cl}

\begin{figure}[ht]
  \centering
\includegraphics[width=\textwidth]{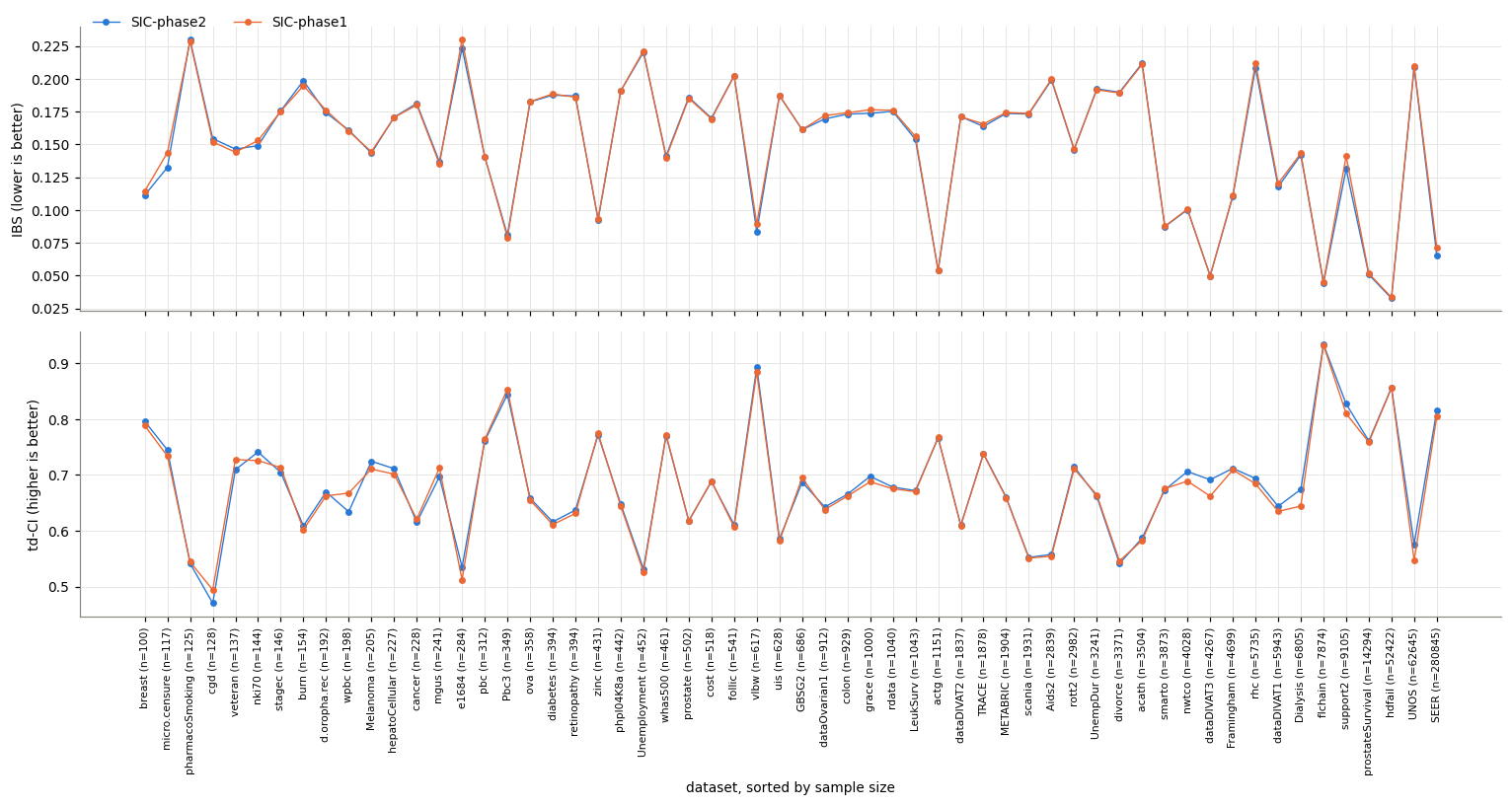}
    \caption{Performance in (upper) Integrated Brier Score (IBS) and (lower) time-dependent C-index (td-CI) across 58 datasets in stage 1 after 5000 steps for different \textit{training phases}. Phase 2 (main option used in evaluation) wins over Phase 1 with win rates: 0.62 (IBS) and 0.71 (td-CI).}
    \label{fig:discretization}
\end{figure}
\newpage
\subsection{Survival Heads}
\label{app:ablation:heads}

\begin{figure}[ht]
  \centering
\includegraphics[width=\textwidth]{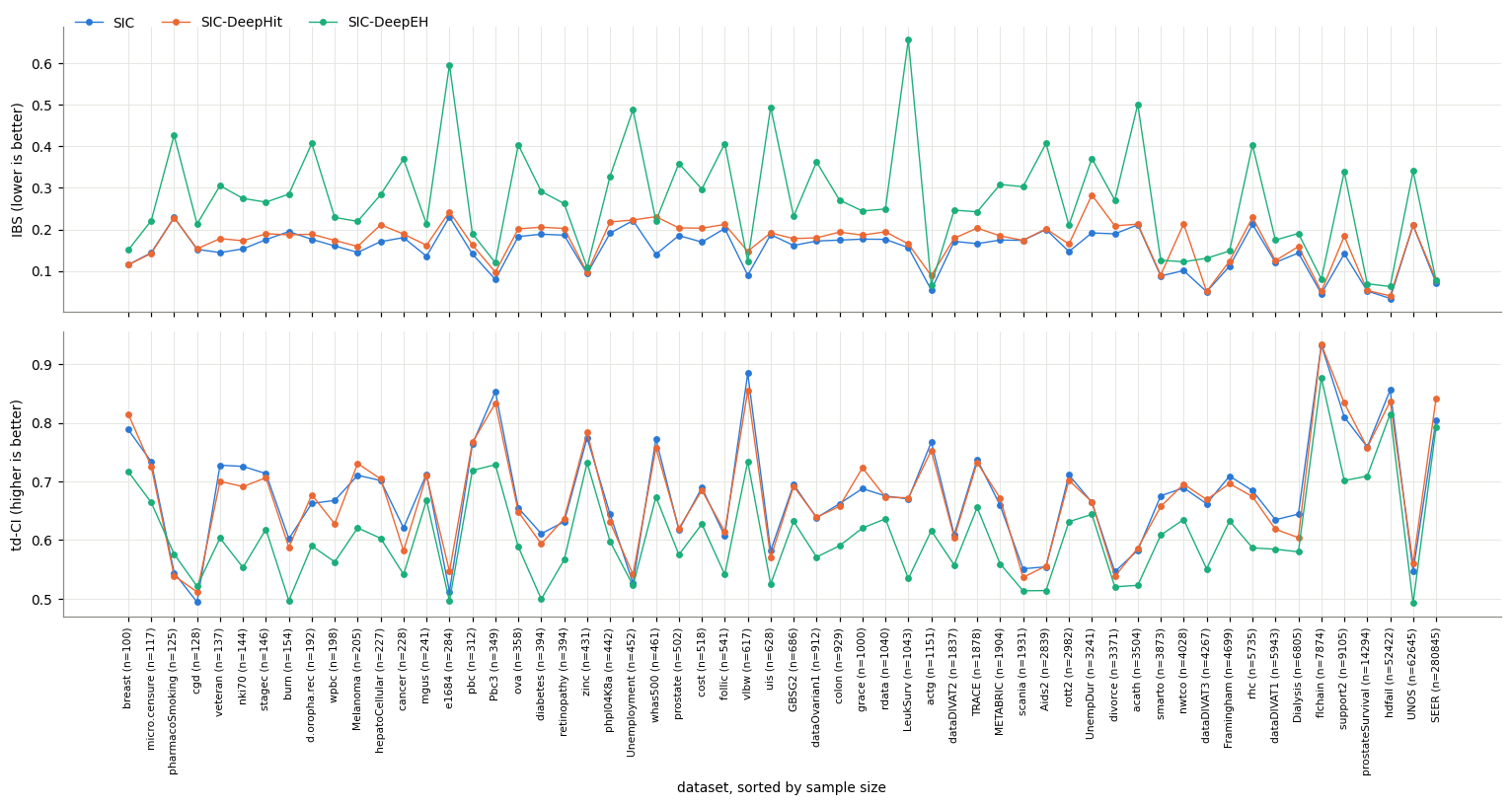}
    \caption{Performance in (upper) Integrated Brier Score (IBS) and (lower) time-dependent C-index (td-CI) across 58 datasets in stage 1 after 5000 steps for different \textit{Survival Heads}.}
    \label{fig:survheads}
\end{figure}

\begin{table}[h]
\centering\small
\caption{Pairwise win-rate comparison on the 58 real-world datasets in stage 1 after 5000 steps for different \textit{Survival Heads}: SIC (SuMo++~\citep{sumopp}), SIC-DeepHit (DeepHit with 10 bins~\citep{lee18}), SIC-DeepEH (Extended Hazard~\citep{zhong21}). Win rate of model A over model B across datasets. }
\label{tab:abl_survival_heads}
\begin{tabular}{llcc}
\toprule
Model A & Model B & td-CI & IBS \\
\midrule
SIC & SIC-DeepHit & 0.57 & 0.94 \\
SIC & SIC-DeepEH & 0.97 & 1.00 \\
\bottomrule
\end{tabular}
\end{table}

\subsection{Informative Censoring and Identifiability}
\label{app:identify}

In the real world, we observe only $(X, T_{obs}, e)$, where $T_{obs}=min(T,C)$ and $e=\mathbf 1\{T \le C\}$, which is derived from marginal event and censoring time distributions and their dependence (with the event-time distribution having the true event-time survival curve $S_T$). Therefore, the observed data identify the distribution of $(T_{obs}, e)$ given $X$, but not its components. Any such distribution can also be generated by a pair $(T, C)$ that is conditionally independent ($T \perp C \mid X$)~\citep{tsiatis75}, with event survival $S^{\#}$. The right-censored likelihood from Equation~\ref{eq:nll} estimates $S^{\#}$. Under non-informative censoring ($T\perp C\mid X$), $S^{\#} = S_T$, so the likelihood is proper objective for $S_T$~\citep{Rindt21}. However, under informative censoring ($T\not\perp C\mid X$), there exist infinitely many $S_T$ that correspond to the same observable law $S^{\#}$, and, thus, $S_T$ cannot be identified from the observed data. Training a PFN on the true event times, which are available during prior generation, removes the misspecification of the loss, but not the non-identifiability

We preliminarily investigate the problem with three models: \textit{SIC} is trained with conditionally independent censoring and the observed-time loss (Equation~\ref{eq:nll}); \textit{SIC (info, true)} is trained on a non-restricted prior ($T$ and $C$ may share $x' \not \in X$, resulting in informative censoring) with true event times (only event term of Equation~\ref{eq:nll}); \textit{SIC (info, obs)} is trained on a non-restricted prior with observed time (Equation~\ref{eq:nll}).

On the same real-world datasets of Section~\ref{sec:exp}, Table~\ref{tab:info_real_pairwise} shows the pairwise win-rates. The SIC  wins the comparison against SIC (info, true) and SIC (info, obs) in td-CI and IBS. This outcome may partly be expected by construction, since td-CI, IBS are calculated under observable law. Under informative censoring,
IPCW-based IBS is not generally proper for $S_T$ and may instead favor $S^{\#}$~\citep{liu26}. Real-world benchmarks can therefore show the cost of a dependence belief, but never its benefit. 

To further investigate the impact on identifiability, we extend our simulated dataset Lognormal-AFT-DAG-1000 with independent censoring (Section~\ref{res:ablation} and Appendix~\ref{res:ablation}) to conditionally independent (\textit{covdep}) and informative (\textit{info}) censoring. Table~\ref{tab:info_sim_obs} shows, in addition to the observed td-CI and IBS, the $\text{surv-}L_2$ ($L_2$ distance to the generator's $S_T$), as a proxy for recovering the true survival function. We observe similar results for IBS and td-CI. Furthermore, on the informative dataset, SIC (info, true) does not come closer to the true event-time $S_T$ in $\text{surv-}L_2$. 

We interpret both results through the posterior predictive that a PFN approximates. With the observed-time loss, the PPD converges to $S^{\#}$ under any prior.
With the true-time loss, the PPD returns the prior average of $S_T$ over the identified set~\citep{moon12}. Our non-restricted prior is primarily dominated by SCMs with conditionally independent censoring and has no preferred direction of dependence, so this average sits near $S^{\#}$. The information needed to identify $S_T$ is not in the context, and providing a survival PFN with true event-time supervision during prior generation is not sufficient to ensure better estimation of the true survival distribution. Restoring identifiability requires assumptions beyond the data. For instance, introducing copulas (e.g., Archimedean~\citep{zhang24}) can restore identifiability under parametric assumptions and let the prior express a controlled range of dependence. We leave the theoretical and empirical study of survival PFNs under informative censoring to future work.

\begin{table}[H]
\centering\small
\caption{Pairwise comparison on the 58 real-world datasets. Win rate of model A over model B across datasets. SIC is trained on a conditionally independent (non-informative) prior with observed time loss. SIC (info, true) is trained on a non-restricted (informative) prior with event true time loss. SIC (info, obs) is trained on a non-restricted (informative) prior with observed time loss.}
\label{tab:info_real_pairwise}
\begin{tabular}{llcc}
\toprule
Model A & Model B & td-CI & IBS\\
\midrule
SIC & SIC (info, true) & 0.63 & 0.74\\
SIC & SIC (info, obs) & 0.53  & 0.63  \\
SIC (info, obs) & SIC (info, true) & 0.61  & 0.69  \\
\bottomrule
\end{tabular}
\end{table}

\begin{table}[H]
\centering\small
\caption{Results for the simulated datasets with independent (indep), conditionally independent (covdep), and informative (info) censoring. td-CI and IBS are scored on the observed time. $\text{surv-}L_2$ is the distance to the true event-time and the predicted survival curve.}
\label{tab:info_sim_obs}
\begin{tabular}{llccc}
\toprule
Dataset & Model & td-CI $\uparrow$ & IBS $\downarrow$ & surv-$L_2$ $\downarrow$ \\
\midrule
Lognormal-AFT-DAG-1000-indep & SIC & 0.793 $\pm$ 0.022 & 0.136 $\pm$ 0.011 & \textbf{0.064 $\pm$ 0.004} \\
 & SIC (info, obs) & \textbf{0.795 $\pm$ 0.018} & \textbf{0.134 $\pm$ 0.010} & 0.069 $\pm$ 0.003 \\
 & SIC (info, true) & 0.786 $\pm$ 0.020 & 0.140 $\pm$ 0.008 & 0.081 $\pm$ 0.008 \\
\midrule
Lognormal-AFT-DAG-1000-covdep & SIC & \textbf{0.785 $\pm$ 0.016} & 0.154 $\pm$ 0.018 & 0.067 $\pm$ 0.003 \\
 & SIC (info, obs) & \textbf{0.785 $\pm$ 0.011} & \textbf{0.153 $\pm$ 0.018} & \textbf{0.065 $\pm$ 0.005} \\
 & SIC (info, true) & 0.776 $\pm$ 0.011 & 0.159 $\pm$ 0.013 & 0.080 $\pm$ 0.006 \\
\midrule
Lognormal-AFT-DAG-1000-info & SIC & \textbf{0.753 $\pm$ 0.024} & \textbf{0.167 $\pm$ 0.014} & \textbf{0.079 $\pm$ 0.003} \\
 & SIC (info, obs) & 0.752 $\pm$ 0.021 & \textbf{0.167 $\pm$ 0.011} & 0.085 $\pm$ 0.003 \\
 & SIC (info, true) & 0.744 $\pm$ 0.024 & 0.171 $\pm$ 0.013 & 0.089 $\pm$ 0.005 \\
\bottomrule
\end{tabular}
\end{table}

\section{Limitations and Outlook.}
\label{app:limitations}
We focus on TabICL as a representative state-of-the-art method due to its scalability and publicly available implementation. Although evaluating multiple prior-fitted models would broaden our comparison, such ablations require substantial computational resources. Future work may extend this analysis to alternative architectures, such as TabDPT~\citep{ma25} or TabPFNv2~\citep{hollmann25}, to assess the generality of our findings under broader model choices. 

Finally, survival analysis is not restricted to right-censoring, a single event of interest, or a single observation time. However, extending our model for left- or interval-censoring~\citep{booksa}, recurrent events~\citep{gupta19}, competing risks~\citep{jeanselme23}, or time-varying effects~\citep{kopper22} would require not only changing the in-context learning architecture itself, but also introducing a new prior formulation that is able to capture these dependencies. 

\section{Broader Impact}
\label{app:impact}
In this work, we present a methodology for survival analysis and further highlight key considerations from the impact statement by~\citet{lee24}. We discuss the direct and indirect implications of our approach through experiments, among others, on real-world clinical datasets, which were used in accordance with the guidelines of the data providers. We acknowledge that novel survival analysis techniques may have both positive and negative impacts. These methods may offer useful insights for personalized treatment, prognosis estimation, and risk stratification. However, their predictions should be interpreted carefully and validated by domain experts before any use in clinical practice. Improper use or over-reliance on survival analysis models could result in unintended harms to individuals or groups, including inequitable access to care or discriminatory outcomes. We therefore stress the importance of transparency, accountability, and sustained dialogue among researchers, clinicians, and the public to promote responsible and ethical use of survival analysis in healthcare.

\end{document}